%% file: main.tex
\documentclass{article}

\usepackage[final]{corl_2026} % Uncomment for the camera-ready ``final'' version.

\usepackage{graphicx}
\usepackage{amsmath}
\usepackage{amssymb}
\usepackage{booktabs}
\usepackage{multirow}
\usepackage{makecell}
\usepackage{caption}
\usepackage{enumitem}
\usepackage{placeins}

\usepackage{titlesec}

\titlespacing*{\section}{0pt}{0.5ex}{0.25ex}
\titlespacing*{\subsection}{0pt}{0.15ex}{0.1ex}
\titlespacing*{\paragraph}{0pt}{0.1ex}{0.2em}
\usepackage[font=small,skip=2pt]{caption}
\title{HARP-VLA: Human-Robot Aligned Representation Learning for Vision-Language-Action Model 
}

\author{
Xiang Zhu$^{1,2*}$ \quad
Puzhen Yuan$^{1*}$ \quad
Yichen Liu$^{1*}$ \quad
Jianyu Chen$^{1,2\dagger}$ \\
$^{1}$ Institute for Interdisciplinary Information Sciences, Tsinghua University, China\\
$^{2}$ Shanghai Qi Zhi Institute, China\\
\texttt{\{zhuxiang24, ypz21, liu-yc22\}@mails.tsinghua.edu.cn} \\
\texttt{jianyuchen@mail.tsinghua.edu.cn} \\
$^{*}$Equal contribution, $^{\dagger}$Corresponding author.
}

\hypersetup{
  pdftitle={HARP-VLA: Human-Robot Aligned Representation Learning for Vision-Language-Action Model},
  pdfauthor={Xiang Zhu, Puzhen Yuan, Yichen Liu, Jianyu Chen}
}

\begin{document}
\maketitle
\vspace{-0.5em}

% 缩小公式间距
\setlength{\abovedisplayskip}{2pt}
\setlength{\belowdisplayskip}{2pt}
\setlength{\abovedisplayshortskip}{2pt}
\setlength{\belowdisplayshortskip}{2pt}

%===============================================================================

\begin{abstract}
Learning generalizable vision-language-action (VLA) models from large-scale human videos is promising but challenging due to cross-embodiment discrepancies in both visual observations and executable actions. 
While latent action models reduce the action execution gap by learning action abstractions, they still rely on visual features. % the visual observation gap still remains.
% Thus, misaligned human and robot representations can therefore induce domain-dependent latent actions and hinder human-robot co-training. 
Thus, misaligned human and robot visual representations can lead to inconsistencies in policy inputs and induce domain-dependent latent actions, hindering effective co-training with human videos.
To address this, we propose HARP, a human-robot aligned representation learning framework for more effective VLA pretraining from human videos. 
Specifically, HARP uses paired human-robot demonstrations as cross-embodiment bridges and abundant unpaired human and robot videos as a scalable dynamics supervision data source. 
It trains a robot-adapted visual encoder and a latent action model with manipulation-centric auxiliary cues and a source-relative pair-discriminative alignment loss, which adapts robot representations toward human semantics while preserving pair-level discrimination.
The learned aligned vision encoder and latent action model provide a unified vision and action representation for VLA-style policy learning, where human and robot videos provide vision-language-to-latent-action supervision and a lightweight robot action head grounds latent actions into executable commands. 
% Experiments across representation visualization and retrieval analysis, simulated benchmarks, and realworld manipulation tasks demonstrate improved human-robot visual and latent-action alignment and consistent gains in downstream manipulation policies. 
% Specifically, we achieve an average length of 4.481 on CALVIN ABC$\rightarrow$D and improve the real-world success rate by 7.1\% compared to the strongest baseline.
Experiments on feature visualization, simulation, and real-world manipulation show improved human-robot alignment and downstream policy performance, 
% achieving 4.481 average length on CALVIN ABC$\rightarrow$D and a 7.1\% realworld success rate gain over the strongest baseline.
achieving 4.481 average length on CALVIN ABC$\rightarrow$D and a 7.1 percentage-point improvement in real-world success rate over the strongest evaluated baseline.
Code, videos, and demos are available at https://puzhenyuan.github.io/HARP-VLA-website/.

\end{abstract}

\keywords{Vision-Language-Action Model, Latent Action Model, Representation Learning, Human Videos} 

%===============================================================================

\section{Introduction}
\input{Intro/intro}

%===============================================================================

\section{Related Works}

% \subsection{Vision-Language-Action Models for Robot Manipulation}

% Vision-Language-Action (VLA) models provide a unified framework for generalizable robot manipulation by jointly modeling visual observations, language instructions, and robot actions with Transformer-based token sequences.

% RT-1 \cite{Brohan2022RT1RT} demonstrated that a single Transformer trained on large-scale realworld demonstrations can generalize across hundreds of manipulation tasks. RT-2 \cite{Brohan2023RT2VM} further showed that web-scale vision-language pretraining improves robotic generalization through semantic knowledge transfer.

% Recent open-source systems, including OpenVLA \cite{Kim2024OpenVLAAO} and Octo \cite{Team2024OctoAO}, adopt pretrained language backbones and scalable visual encoders for cross-task and cross-embodiment learning. RDT-1B \cite{Liu2024RDT1BAD} and UniVLA \cite{Bu2025UniVLALT} further address heterogeneous action spaces, while $\pi_0$~\cite{black2026pi0visionlanguageactionflowmodel} and $\pi_{0.5}$ \cite{intelligence2025pi05visionlanguageactionmodelopenworld} extend VLA models to continuous action generation with large-scale multimodal training.

% Together, these works indicate a clear shift toward unified architectures, multimodal pretraining, and embodiment-agnostic policy learning.
% \subsection{Vision-Language-Action Models for Robot Manipulation}

\textbf{Vision-Language-Action Models for Robot Manipulation.} Vision-Language-Action (VLA) models provide a unified framework for robot manipulation by modeling visual observations, language instructions, and actions as Transformer-based token sequences. RT-1~\cite{Brohan2022RT1RT} and RT-2~\cite{Brohan2023RT2VM} show that large-scale robot demonstrations and web-scale vision-language pretraining improve cross-task generalization and semantic reasoning. Recent open-source VLAs further scale this paradigm toward multi-task, cross-embodiment, and continuous-action policy learning. OpenVLA~\cite{Kim2024OpenVLAAO} and Octo~\cite{Team2024OctoAO} leverage pretrained language and visual backbones, while RDT-1B~\cite{Liu2024RDT1BAD}, UniVLA~\cite{Bu2025UniVLALT}, $\pi_0$~\cite{black2026pi0visionlanguageactionflowmodel}, and $\pi_{0.5}$~\cite{intelligence2025pi05visionlanguageactionmodelopenworld} extend VLA models to heterogeneous action spaces and continuous action generation.
\textbf{Learning from Human Videos via Cross-Embodiment Alignment.} 
Human videos provide demonstrations for robot learning, but are limited by embodiment gaps. Existing methods address cross-embodiment gaps through trajectory cues, object-motion modeling, or representation alignment.
RT-Trajectory~\cite{Gu2023RTTrajectoryRT}, MimicPlay~\cite{Wang2023MimicPlayLI}, and Learning by Watching~\cite{Xiong2021LearningBW} use hand keypoint trajectories to guide policies.
Object-centric methods such as Im2Flow2Act~\cite{Xu2024FlowAT} and Dream2Flow~\cite{dharmarajan2025dream2flow} model object motion as an embodiment-invariant signal. 
% whereas Human2Robot~\cite{Xie2025Human2RobotLR} and Gen2Act~\cite{Bharadhwaj2024Gen2ActHV} synthesize robot demonstrations.
More recent work reduces the human--robot domain gap through visual transformation or representation alignment.
EgoMimic~\cite{Kareer2024EgoMimicSI} uses visual alignment cues, RoVi-Aug~\cite{Chen2024RoViAugRA} and DexUMI~\cite{Xu2025DexUMIUH} generate embodiment-consistent observations, and HR-Align~\cite{Zhou2024MitigatingTH} learns shared feature spaces with contrastive objectives.
These studies show that human videos can benefit robot manipulation when cross-embodiment variation is reduced through object or motion cues, visual observations, or learned representations.
% \subsection{Latent Action Models and Cross-Embodiment Alignment}
% Since human videos usually lack robot-executable action labels, recent studies learn embodiment-agnostic latent actions from unlabeled or weakly labeled data. Instead of predicting robot commands directly, these methods construct shared motion embeddings that support cross-embodiment transfer through a unified action space.

% LAPA \cite{Ye2024LatentAP} learns discrete latent action codes from large-scale unlabeled videos, aligns them with vision-language representations, and maps them to robot controls using limited robot data. UniVLA \cite{Bu2025UniVLALT} improves this formulation by emphasizing task-relevant motion and suppressing irrelevant dynamics, thereby enhancing multi-source pretraining and cross-embodiment generalization. UniSkill \cite{Kim2025UniSkillIH} and IGOR \cite{Chen2024IGORIR} further use inverse and forward dynamics objectives to learn motion-centric and semantically consistent embeddings shared by humans and robots.

% Overall, latent action modeling provides a scalable interface between human videos and robot manipulation policies.
% \subsection{Latent Action Models and Cross-Embodiment Alignment}

\textbf{Latent Action Models for Cross-Embodiment Policy Learning.} Since human videos usually lack executable action labels, recent studies learn embodiment-agnostic latent actions from unlabeled or weakly labeled data. Rather than predicting robot commands, they learn shared motion embeddings as a unified action space for cross-embodiment transfer. LAPA~\cite{Ye2024LatentAP} learns discrete latent action codes from large-scale unlabeled videos and maps them to robot controls with limited robot data, while UniVLA~\cite{Bu2025UniVLALT} improves this formulation by emphasizing task-relevant motion and suppressing irrelevant dynamics. UniSkill~\cite{Kim2025UniSkillIH} and IGOR~\cite{Chen2024IGORIR} further introduce inverse and forward dynamics objectives to learn motion-centric embeddings shared by humans and robots. Overall, latent action modeling provides a scalable interface between human videos and robot manipulation policies.

%===============================================================================

\section{Methods}

\input{Methods/data_process}

\begin{figure}[t]
  \centering  
  \includegraphics[scale=1.0,width=0.99\linewidth]{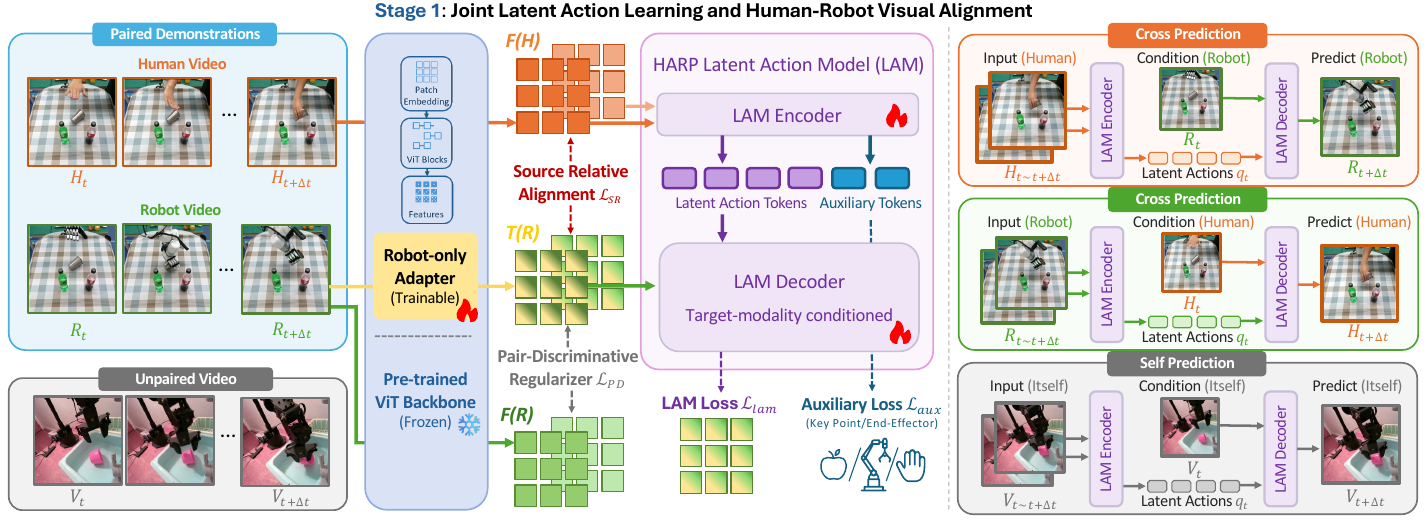} 
\caption{
Stage 1: \textbf{Joint Visual and Latent-Action Alignment}. 
\textbf{Left:} Human to robot cross-prediction example.
HARP jointly learns a robot-adapted visual encoder and a latent action model using paired videos for cross-prediction and alignment, and unpaired videos for self-prediction. 
Auxiliary cues guide latent-action learning. Source-Relative (SR) loss \(\mathcal{L}_{\mathrm{SR}}\) aligns robot features toward humans, while Pair-Discriminative (PD) regularizer \(\mathcal{L}_{\mathrm{PD}}\) preserves representation structure.
\textbf{Right:} Examples of cross-prediction and self-prediction.
}
  \label{fig:stage1}
\end{figure}

\input{Methods/ve_lam_joint_training_new}

\subsection{VLA Pretraining with Aligned Representations}
\label{sec:pretrain}

\begin{figure}[t]
  \centering  
  \includegraphics[scale=1.0,width=0.99\linewidth]{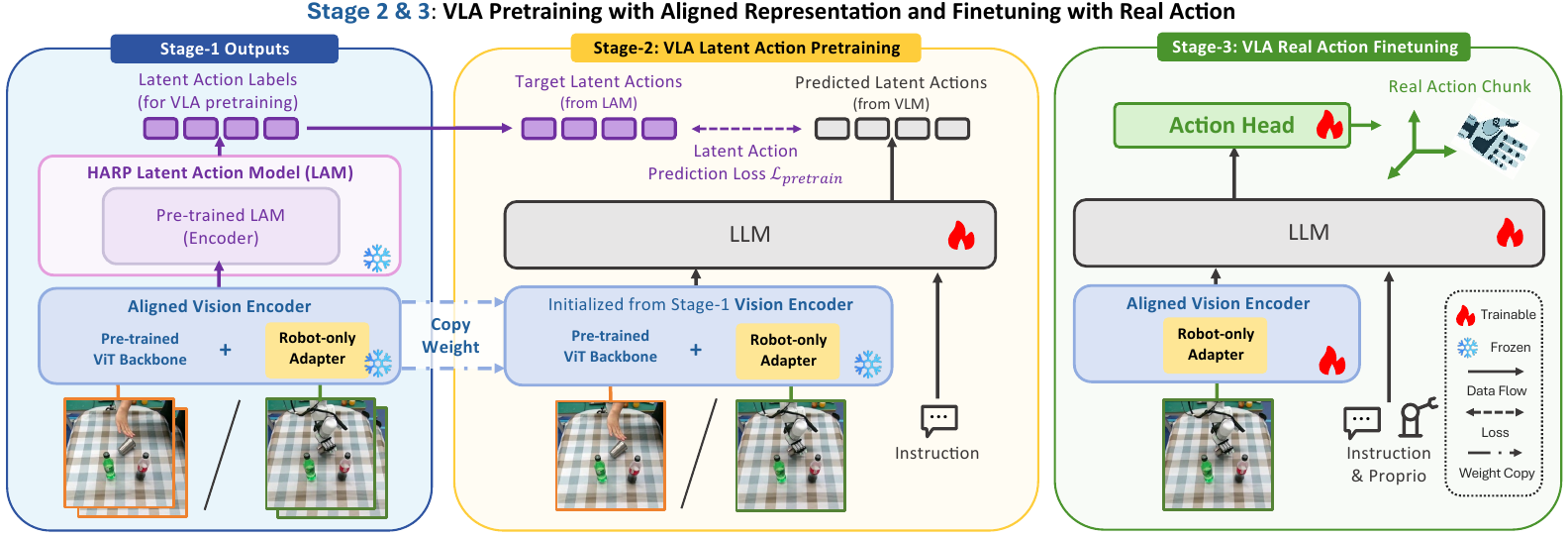} 
  \caption{
\textbf{Pretrain and finetune.} Stage-2: \textbf{Pretrain}. The Stage-1 LAM is used to produce human-robot aligned latent actions, and the aligned vision encoder is used to replace the VLM vision encoder. Stage-3: \textbf{Finetune}. A trainable action head is employed to convert latent action embeddings to executable real actions.
}
  \label{fig:pretrain_finetune}
\end{figure}

Leveraging the latent action model and aligned vision encoder trained in stage-1, we can label video frame $x^t \in X$ with unified latent action $q_X^t = Q_\theta(E_\theta(z_X^t, z_X^{t+\Delta t}, l_X))$, where $z_X^t=\Phi_\theta(x^t, e_X)$, $z_X^{t+\Delta t}=\Phi_\theta(x^{t+\Delta t}, e_X)$, which can be employed to train a generalist policy, as shown in Fig.~\ref{fig:pretrain_finetune}.

We use latent action labels to pretrain a VLM into a VLA,  which uses the same vision encoder as HARP-LAM. The architecture details can be found in the Appendix~\ref{app:model}.
Specifically, the model $\pi_\theta$ receives language instruction $l_X$ and visual input $x^t$, and outputs latent action tokens $\{\hat{q}_i\}_{i=1}^{N_q}$, and is optimized to minimize the cross-entropy loss 
\[
\mathcal{L}_{pretrain}=-\mathbb{E}_{(x^t,l_X)\sim \mathcal{D}}
\left [
\textstyle \sum_{i=1}^{N_q} \log \pi_\theta(\hat{q}_i=q_{X,i}^t|x^t,l_X) 
\right ],
\]
where $N_q$ represents the number of latent action tokens.
This approach leverages the original visual reasoning and language understanding capabilities of the pretrained VLM, while additionally equipping it with action modeling ability by training it to predict latent action indices. 

% TODO：视效果来看是不是用lam的vit的权重会好一点
In addition to directly pretraining on original VLM, we further propose copying the adapter weights learned in the stage-1 (\ref{sec:lam}) to the vision encoder of VLM, as shown in Fig.~\ref{fig:pretrain_finetune}. 
% Considering that the VLM is pre-trained on large-scale internet data, the vision encoder tends to model the human and robot features independently. 
In this way, we enable the vision encoder to achieve human–robot visual alignment as well as latent action alignment, facilitating effective pre-training on large-scale human videos for robot manipulation.

\subsection{VLA Finetuning with Real Action}
\label{sec:finetune}

As the VLA is pretrained with latent actions, it cannot directly output real actions for control. To address this, we finetune it with a small amount of real-action robot data for downstream deployment.

Specifically, we employ an action head that maps from the latent action embedding to normalized real action. 
We train the action head from scratch using L1 loss, and simultaneously use LoRA~\cite{hu2022lora} to finetune the VLA backbone to achieve effective adaptation.
We broadly refer to the policy model that has gone through these stages as HARP-VLA, with architecture details in Appendix~\ref{app:model}.

%===============================================================================

\section{Experiments}
In stage-1, we train HARP-LAM on human datasets  OpenEgo~\cite{jawaid2025openego}, robot datasets Bridge-V2~\cite{walke2023bridgedata}, and human-robot paired datasets RH20T~\cite{fang2023rh20t}, Human2Robot~\cite{Xie2025Human2RobotLR} and a self-collected paired dataset on Robotera XHand. 
In stage-2, we pretrain our VLA on the same datasets, using only video data and latent action labels extracted from HARP-LAM.
In the last stage, we further fine-tune the HARP-VLA according to downstream tasks: CALVIN~\cite{mees2022calvin}, and real-world experiment.

The goal of our experiments is to validate the effectiveness of the proposed HARP framework and test HARP-VLA's general control ability.
We evaluate HARP along five aspects: 
(1) human-robot visual and latent-action alignment, 
(2) component contributions, 
(3) downstream policy utility of the aligned visual encoder, 
(4) full HARP-VLA policy performance, and 
(5) robustness to reduced paired-data diversity.

\begin{figure}[t]
  \centering  
  \includegraphics[scale=1.0,width=0.99\linewidth]{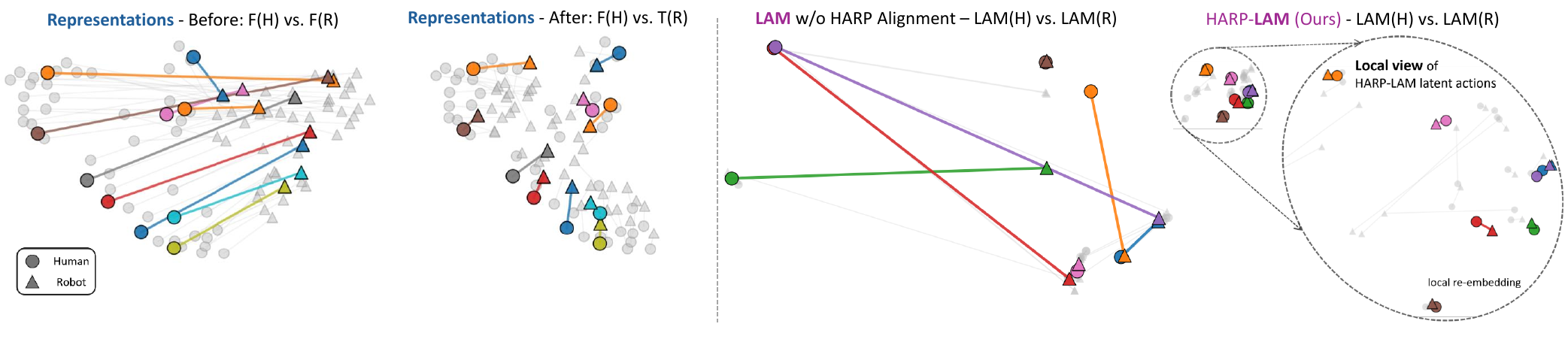} 
  \caption{
UMAP visualization of human-robot alignment. 
Left: visual representations before adaptation, $F(H)$ vs.~$F(R)$, and after HARP adaptation, $F(H)$ vs.~$T(R)$. 
Right: latent-actions without and with HARP-LAM alignment. 
Circles and triangles denote human and robot; colored pairs are highlighted for readability.
}
\label{fig:umap}
\end{figure}

\subsection{Human-Robot Representation and Latent-Action Alignment}
\label{sec:alignment}

We first evaluate whether HARP reduces the human-robot representation gap before turning to downstream policy learning. 
% We use three complementary evaluations: (1) UMAP visualization for qualitative inspection, (2) paired human-robot cosine distance for direct geometric alignment, and (3) bidirectional cross-embodiment retrieval for semantic discriminability.
% 【Update for Camera-Ready】：增加额外的Quantitative LAM exp 
We use complementary qualitative and quantitative evaluations, including UMAP visualization, paired human-robot cosine distance, cross-embodiment visual retrieval, and latent-action correspondence.

\paragraph{Baselines.}
We compare against the unadapted vision encoder and HR-Align-based~\cite{Zhou2024MitigatingTH} baselines. 
HR-Align (denoted as HR) uses its original task-aware pooling design, where the language instruction attends to video features before contrastive alignment. 
HR-Style removes task-aware pooling and applies the same contrastive loss directly on pooled features.
We also evaluate HARP variants by changing alignment objective and pooling space. 
HARP-HR uses original HR-Align objective with task-aware pooling, while HARP-HR-Style applies HR-Align contrastive loss directly on pooled HARP features. 
HARP-L2 uses a naive L2 loss as the alignment loss on paired features.
HARP-SR uses only the source-relative term, HARP-SRPD further adds the pair-discriminative term. 
% and HARP-SRPD+RP additionally includes the relational preservation loss.

\paragraph{Comparison protocol.} All HARP variants use the same backbone, robot-only adapter, data split and preprocess, and training budget; only the alignment objective or pooling space differs.
HR-Align baselines use the same fused backbone and adapter but follow their original paired-data alignment setup.
All representation metrics are evaluated on the same held-out paired videos with identical sampling, pooling, and cosine similarity. 
See Appendix~\ref{app:align_eval} for details.

\paragraph{Qualitative visualization.}
Fig.~\ref{fig:umap} provides a qualitative comparison of the learned visual representation and latent-action spaces. For visual representations, we compare the original feature space before adaptation, i.e., $F(H)$ versus $F(R)$, with the adapted feature space after HARP alignment, i.e., $F(H)$ versus $T(R)$. Before adaptation, human and robot demonstrations form visibly separated clusters, and even semantically matched pairs often remain far apart. After adaptation, paired human-robot samples become substantially closer, with a noticeably reduced cross-embodiment gap. A similar trend is observed in the latent-action space. Without HARP alignment, latent actions extracted from human and robot videos are dispersed and often poorly matched across embodiments. In contrast, HARP-LAM produces a more compact and better aligned latent-action space, where paired human and robot motions are mapped to nearby regions. These qualitative results suggest that HARP improves alignment at both the visual representation level and the latent-action level.

\begin{figure}[t]
\centering

\begin{minipage}[t]{0.46\linewidth}
    \vspace{0pt}
    \centering
    \resizebox{\linewidth}{!}{
    \begin{tabular}{lccc}
    \toprule
    Method 
    & H2R R@1 $\uparrow$ 
    & R2H R@1 $\uparrow$ 
    & Avg. R@1 $\uparrow$ \\
    \midrule
    Unadapted & 44.09 & 43.01 & 43.55 \\
    HR & 45.16 & 45.16 & 45.16 \\
    HARP-HR & 46.24 & 60.22 & 53.23 \\
    HARP-L2 & 70.97 & 52.69 & 61.83 \\
    HARP-SR & 84.95 & 64.52 & 74.74 \\
    HARP-SRPD & \textbf{87.10} & \textbf{69.89} & \textbf{78.50} \\
    \bottomrule
    \end{tabular}
    }
    \captionof{table}{
    Bidirectional cross-embodiment retrieval on held-out paired demonstrations.
    H2R uses human videos as queries to retrieve paired robot videos, while R2H retrieves human from robot.
    }
    \label{tab:retrieval}
\end{minipage}
\hfill
\begin{minipage}[t]{0.48\linewidth}
    \vspace{0pt}
    \centering
    \includegraphics[width=\linewidth]{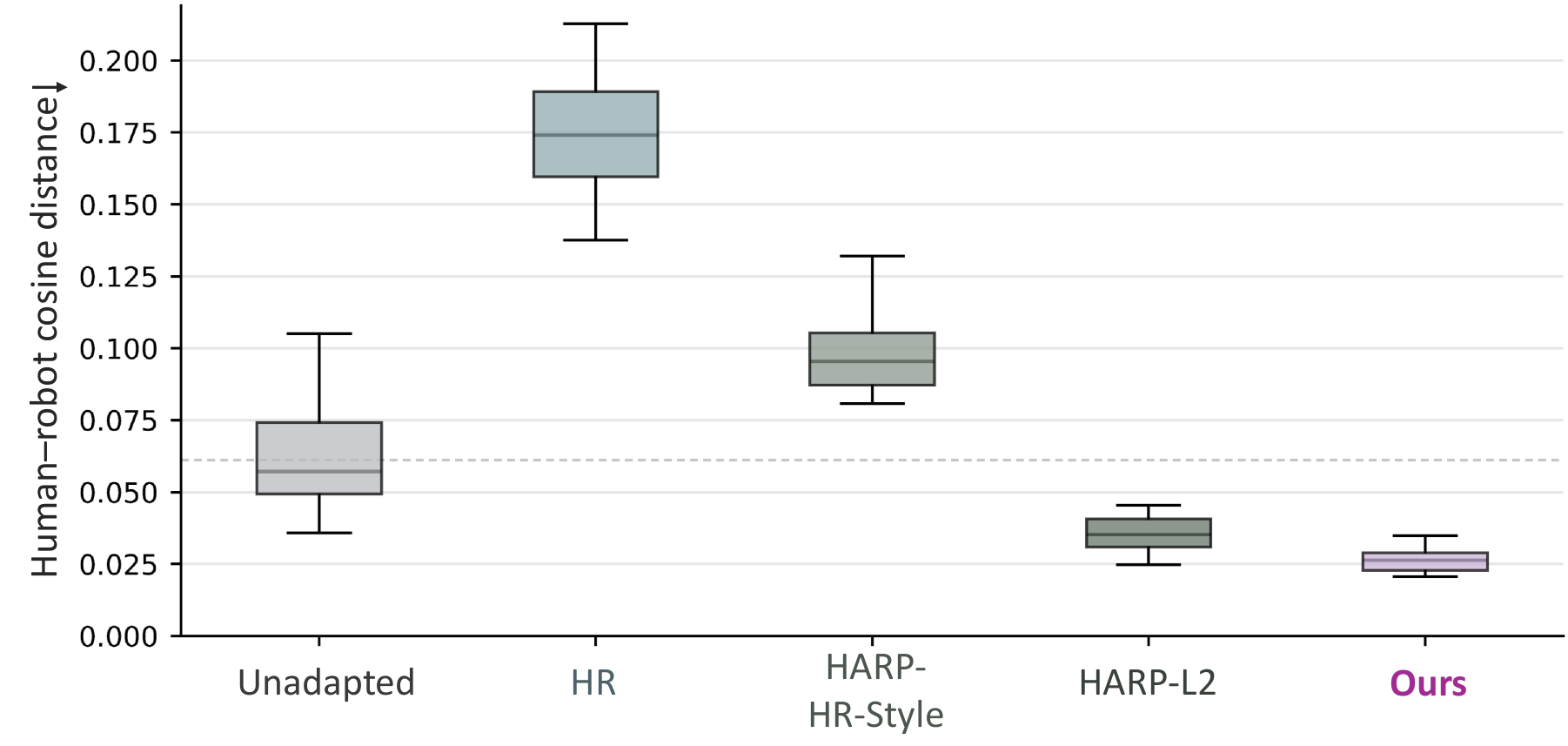}
    \captionof{figure}{Paired human-robot cosine distance.}
    \label{fig:paired_distance}
\end{minipage}

\end{figure}

\paragraph{Paired human-robot cosine distance.}
We further measure geometric alignment using paired cosine distance. 
For unadapted encoder, we measure 
\(d^{\mathrm{orig}}=1-\cos(\rho(F(H)),\rho(F(R)))\). 
For adapted methods, we keep the human branch frozen and only adapt the robot branch, measuring 
\(d^{\mathrm{adapt}}=1-\cos(\rho(F(H)),\rho(T_\theta(R)))\).
% Here, \(T_\theta\) denotes the robot-adapted encoder, and the human branch is never passed through the adapter. 
As shown in Fig.~\ref{fig:paired_distance}, HR-Align-based objectives do not reduce the paired distance 
% below the unadapted encoder 
when evaluated in the same mean-pooled visual representation space. 
% In contrast, our methods substantially decreases the distance, showing that our source-relative pair-discriminative alignment directly mitigates the absolute human-robot representation gap.
In contrast, HARP variants substantially reduce the paired distance in the evaluation space, indicating that the learned robot-adapted features are geometrically closer to their paired human counterparts.

\paragraph{Cross-embodiment retrieval.}

Paired distance measures absolute geometric alignment, but it does not evaluate whether the learned space remains discriminative across different demonstrations. 
We therefore perform cross-embodiment retrieval on held-out paired videos. 
Given a human query, H2R retrieval ranks all robot videos by cosine similarity and checks whether the paired robot is retrieved; R2H retrieval is defined symmetrically. 
We report Recall@1. As shown in Table~\ref{tab:retrieval}, HARP-SRPD improves from 43.55 to 78.50 in averaged R@1 over the unadapted encoder and outperforms HR-Align by a large margin. 
Compared with HARP-SR, adding the pair discriminative term further improves retrieval, indicating stronger cross-embodiment discriminability.

\paragraph{Cross-embodiment latent-action correspondence.} 
We further evaluate alignment in the latent-action space by applying the same bidirectional retrieval protocol to transition-level latent-action embeddings, reporting the averaged H2R/R2H Recall@1 as Lat. R@1. We additionally report the code-agreement gap (CA-gap) in percentage points, which measures how often temporally corresponding human-robot transitions share discrete VQ codes relative to shuffled pairs. As shown in Table~\ref{tab:component_ablation}, the latent retrieval and code-agreement results provide quantitative evidence of consistent cross-embodiment latent-action correspondence.

\begin{table}[h]
    \centering
    \setlength{\tabcolsep}{5pt}
    \renewcommand{\arraystretch}{0.95}
    \small
    \begin{tabular}{lcccccc}
        \toprule
        \textbf{Variant} &
        \textbf{Pair Pred.} &
        \textbf{Unpair.} &
        \textbf{Aux} &
        \textbf{Vis. R@1$\uparrow$} &
        \textbf{Lat. R@1$\uparrow$} &
        \textbf{CA-gap(pp)$\uparrow$} \\
        \midrule
        Full HARP   & S+C & Yes & Yes & \textbf{78.50} & \textbf{36.29} & \textbf{30.67} \\
        Self-only   & S   & Yes & Yes & 73.93 & 26.62 & 22.24 \\
        Cross-only  & C   & Yes & Yes & 75.10 & 28.03 & 25.16 \\
        Paired-only & S+C & No  & Yes & 75.16 & 32.06 & 30.04 \\
        w/o Aux     & S+C & Yes & No  & 73.37 & 34.07 & 26.71 \\
        \bottomrule
    \end{tabular}
    \caption{
    Controlled Stage-1 component ablations.
    S/C denote self-/cross-prediction on paired samples;
    unpaired samples use self-prediction. Full HARP denotes HARP-SRPD with self- and cross-prediction, unpaired data, and auxiliary supervision.
    }
    \label{tab:component_ablation}
    \vspace{-4pt}
\end{table}

%【update for camera-ready】
\subsection{Component Ablations}
As shown in Table~\ref{tab:component_ablation}, combining self- and cross-prediction consistently improves both visual and latent-action correspondence over either objective alone. Removing unpaired videos or auxiliary supervision also degrades alignment, indicating that these components provide complementary supervision.

\subsection{Robot Policy Learning with Aligned Visual Representations}
\label{sec:rlbench}
We next evaluate whether the aligned visual representation is useful for downstream robot policy learning. 
To evaluate the ability of HARP-VE for downstream manipulation policy learning, we follow HR-Align \cite{Zhou2024MitigatingTH} to evaluate different methods on RLBench \cite{james2020rlbench}, which is challenging for its multi-task setting with significant task diversity and difficulty. All methods use the same policy architecture, training data, action space, and optimization budget. 
Thus, differences in success rate mainly reflect the quality of the learned visual representation.

As shown in Table~\ref{tab:rlbench_repr}, HARP-SRPD achieves the best policy performance, improving the average success rate from 37.56 to 46.59 over the unadapted encoder. 
Compared with HARP-SR, adding the pair-discriminative term further improves downstream success, suggesting that stronger cross-embodiment discriminability translates into more useful robot features.

\begin{figure}[h]
\centering

\begin{minipage}[t]{0.33\linewidth}
    \vspace{0pt}
    \centering
    \resizebox{\linewidth}{!}{
    \begin{tabular}{lc}
    \toprule
    Method & Avg. $\uparrow$ \\
    \midrule
    Unadapted & 37.56 \\
    HR & 39.70 \\
    HR-Style & 38.22 \\
    HARP-HR & 35.11 \\
    HARP-HR-Style & 40.07 \\
    HARP-L2 & 40.78 \\
    HARP-SR & 43.41 \\
    HARP-SRPD & \textbf{46.59} \\
    \bottomrule
    \end{tabular}
    }
    \captionof{table}{Average success rate (\%) on 18 RLBench tasks using frozen visual encoders.}
    \label{tab:rlbench_repr}
\end{minipage}
\hfill
\begin{minipage}[t]{0.64\linewidth}
    \vspace{0pt}
    \centering
    % 关键：让右边这个 caption 显示为 Table 4
    % \setcounter{table}{3}

    \setlength{\tabcolsep}{2.5pt}
    \resizebox{\linewidth}{!}{
    \begin{tabular}{lccccc}
    \toprule
    Model 
    & Pick $\uparrow$
    & Push $\uparrow$
    & Press $\uparrow$
    & Flip $\uparrow$
    & Avg. $\uparrow$ \\
    \midrule
    $\pi_0$~\cite{black2026pi0visionlanguageactionflowmodel} 
    & 58.3 & 75.0 & 56.7 & 35.0 & 56.3 \\
    $\pi_{0.5}$~\cite{intelligence2025pi05visionlanguageactionmodelopenworld}
    & 71.7 & \textbf{83.3} & 68.3 & 53.3 & 69.2 \\
    OpenVLA~\cite{Kim2024OpenVLAAO}
    & 0.0 & 23.3 & 18.3 & 0.0 & 10.4 \\
    UniVLA~\cite{Bu2025UniVLALT} 
    & 38.3 & 61.7 & 31.7 & 21.7 & 38.4 \\
    OpenVLA-OFT~\cite{kim2025fine} 
    & 51.7 & 71.7 & 76.7 & 43.3 & 60.9 \\
    \midrule
    HARP-VLA (L2)
    & 70.0 & 71.7 & 81.7 & 56.7 & 70.0 \\
    HARP-VLA (w/o F.)
    & \textbf{76.7} & 80.0 & 78.3 & 58.3 & 73.3 \\
    HARP-VLA (Ours)
    & \textbf{76.7} & 81.7 & \textbf{85.0} & \textbf{61.7} & \textbf{76.3} \\
    \bottomrule
    \end{tabular}
    }
    \captionof{table}{
    Real-world manipulation success rate (\%) averaged over 60 trials per task.
    L2 indicates using L2 alignment loss instead of SRPD.
    F. indicates freezing vision encoder during VLA pretraining.
    }
    \label{tab:realworld}
\end{minipage}
\end{figure}

\subsection{HARP-VLA Policy Evaluation}
\label{sec:harpvla}

\noindent\textbf{Experiment Setup.} 
% 是否需要画图，描述数据集分布？
% The training pipeline consists of three stages: joint visual and latent action alignment, VLA pretraining with aligned representation, and VLA finetuning with real action. 
% We pretrain our HARP-VLA to predict aligned latent actions, and finetune it to execute tasks CALVIN \cite{mees2022calvin} and real-world.
We pretrain HARP-VLA to predict aligned latent actions and finetune it for downstream control on CALVIN \cite{mees2022calvin} and our real-world tasks.
Specifically, we choose the most challenging CALVIN ABC$\rightarrow$D settings in simulation, and finetune it to control an XArm7 with Robotera XHand in real-world.
Across all experimental settings, HARP-VLA takes third-view and wrist camera images, task instructions, and proprioceptive states as input, and outputs a 10-step real-action chunk.
Further simulation and real-world details are in Appendix~\ref{app:exp}.

\textbf{Baselines.} We compare HARP-VLA against 5 representative baselines, including OpenVLA, UniVLA, OpenVLA-OFT, $\pi_0$, and $\pi_{0.5}$, where UniVLA is similar to our training method, and OpenVLA-OFT is similar to our architecture.
We also evaluate HARP  with L2 alignment loss and without vision encoder freezing in pretraining as ablations.
Baseline details are in Appendix~\ref{app:exp}.

Due to differences in backbone, pretraining data, and visual input configurations, these results should be interpreted as system-level comparisons rather than fully controlled architectural comparisons.

\begin{table}[t]
\centering
\caption{
Success rate (\%) and average length on CALVIN benchmark. L2 indicates using L2 alignment loss instead of SRPD. F. indicates freezing the vision encoder in VLA pretraining.
}
\label{tab:simulation}
\begin{tabular}{lcccccc}
\toprule

\addlinespace[-1pt]

Model 
& Task1 $\uparrow$
& Task2 $\uparrow$
& Task3 $\uparrow$
& Task4 $\uparrow$
& Task5 $\uparrow$
& Avg. Len.$\uparrow$ \\
\addlinespace[-1pt]

\midrule

\addlinespace[-1pt]

$\pi_0$~\cite{black2026pi0visionlanguageactionflowmodel}
& 92.3 
& 82.4 
& 72.1 
& 62.2 
& 53.7 
& 3.627 \\

\addlinespace[-1pt]

$\pi_{0.5}$~\cite{intelligence2025pi05visionlanguageactionmodelopenworld}
& 94.4 
& 86.0 
& 76.4 
& 69.7 
& 61.0 
& 3.875 \\

\addlinespace[-1pt]

OpenVLA~\cite{Kim2024OpenVLAAO}
& 91.3 
& 77.8 
& 62.0 
& 52.1 
& 43.5 
& 3.270 \\

\addlinespace[-1pt]

UniVLA~\cite{Bu2025UniVLALT}
& 95.4 
& 85.5 
& 75.4 
& 66.9 
& 56.5 
& 3.800 \\

\addlinespace[-1pt]

OpenVLA-OFT~\cite{kim2025fine}
& 94.2 
& 86.4 
& 78.0 
& 70.4 
& 62.7 
& 3.917 \\

\addlinespace[-1pt]

\midrule

\addlinespace[-1pt]

HARP-VLA (L2) 
& 95.8 
& 89.7 
& 81.3 
& 72.8 
& 64.8 
& 4.044 \\

\addlinespace[-1pt]

HARP-VLA (w/o F.) 
& 98.8 
& 93.9 
& 86.1 
& 77.7 
& 68.5 
& 4.250 \\

\addlinespace[-1pt]

HARP-VLA (Ours) 
& \textbf{99.8} 
& \textbf{96.7} 
& \textbf{91.3} 
& \textbf{84.4} 
& \textbf{75.9} 
& \textbf{4.481} \\

\addlinespace[-1pt]

\bottomrule
\end{tabular}
\end{table}

\textbf{Experiment Results.}
Table~\ref{tab:realworld} and Table~\ref{tab:simulation} show that HARP-VLA achieves an average length of 4.481 on CALVIN ABC$\rightarrow$D and a mean success rate of 76.3\% on the real-world tasks.

% The results presented at Tab.~\ref{tab:realworld} and Tab.~\ref{tab:simulation} demonstrate the advantages of HARP-VLA compared to other baseline methods in simulation and realworld experiments, which improves calvin average length to 4.481, and improves realworld mean success rate to 76.3\%.
% It's worth noting that OpenVLA-OFT and HARP-VLA use similar architectures, the difference being that the former leverages OpenVLA pretrained weights, while the latter uses our HARP method for pretraining, during which no real actions are seen.

% TODO: 加一点具体数值的分析

In ablations, we compare SRPD with L2 alignment loss and study vision encoder freezing during pretraining, as shown in Table~\ref{tab:realworld} and Table~\ref{tab:simulation}.
Although the simple L2 alignment loss achieves competitive feature alignment and RLBench performance, SRPD performs better in VLA pretraining, improving performance by 0.437 on CALVIN average length and 6.3 percentage points in real-world success rate.
% Freezing the vision encoder during pretraining also improves performance by 0.231 on CALVIN average length and 3.0\% in realworld success rate, suggesting that the human-robot vision gap maybe one of the reasons why the vision encoder is typically unfrozen when adapting a VLM into a VLA, as VLMs are primarily pretrained on more human-centric image and video data. 
% By using a robot-only adapter to bridge this gap and freezing the vision encoder in pretraining, the features learned during web-scale pretraining can be better preserved and utilized in robot manipulation tasks.
% Freezing the vision encoder during pretraining improves CALVIN average length by 0.231 and real-world success rate by 3.0\%, suggesting that the human-robot vision gap may partly explains why VLM-to-VLA adaptation typically unfreezes the vision encoder. 
%  Using a robot-only adapter to bridge this gap enables frozen pretraining to better preserve and utilize web-scale visual features for robot manipulation.
Freezing the HARP-initialized vision encoder during Stage-2 pretraining further improves CALVIN average length from 4.250 to 4.481 and real-world success rate from 73.3\% to 76.3\%.
This suggests that, after the robot-only adapter bridges the human-robot visual gap, freezing the vision encoder during latent-action pretraining helps preserve both the learned human-robot alignment and the web-scale pretrained visual features for robot manipulation.

\subsection{Paired-Data Scaling} 
Finally, we investigate how HARP scales with the diversity of unique paired demonstrations. Fig.~\ref{fig:paired_data_scaling} reports visual and latent-action retrieval across five settings, including an unpaired-only baseline (0\%), while Table~\ref{tab:paired_scaling} additionally evaluates RLBench and CALVIN performance at 100\%, 25\%, and 5\%. 
Even 1\% of paired data substantially improves cross-embodiment alignment over unpaired-only training, with visual retrieval showing greater robustness than latent-action retrieval as paired diversity decreases. 
Moreover, HARP maintains strong downstream performance with only 5\% of unique paired demonstrations, suggesting that a substantially smaller set of paired bridges can still support effective alignment and policy transfer when paired-supervision exposure is maintained. 
Additional details are provided in Appendix~\ref{app:ablation_scale}.

% Finally, we study how HARP scales with the amount of paired training data. We uniformly retain 25\%, 5\%, 1\% or 0\% of the unique samples in each paired dataset while keeping the dataset-mixture probabilities, unpaired data, and optimization budget fixed. 
% % Consequently, the remaining paired samples are revisited more frequently, while the overall exposure to paired supervision remains approximately unchanged. This setting isolates the effect of reducing unique paired-data diversity rather than the number of paired-supervision updates.
% Figure~\ref{fig:paired_data_scaling} and Table~\ref{tab:paired_scaling} reports representation alignment, frozen-encoder policy learning, and downstream VLA performance under progressively reduced paired data. 
% HARP degrades gracefully across all stages even with substantially fewer unique paired demonstrations, indicating that a smaller set of paired bridges can still support effective alignment and downstream transfer when training exposure is maintained.
% Additional experimental details are provided in Appendix~\ref{app:ablation_scale}. 

\noindent\makebox[\linewidth][c]{%
\begin{minipage}[t]{0.37\linewidth}
    \vspace{0pt}
    \centering
    \includegraphics[width=\linewidth]{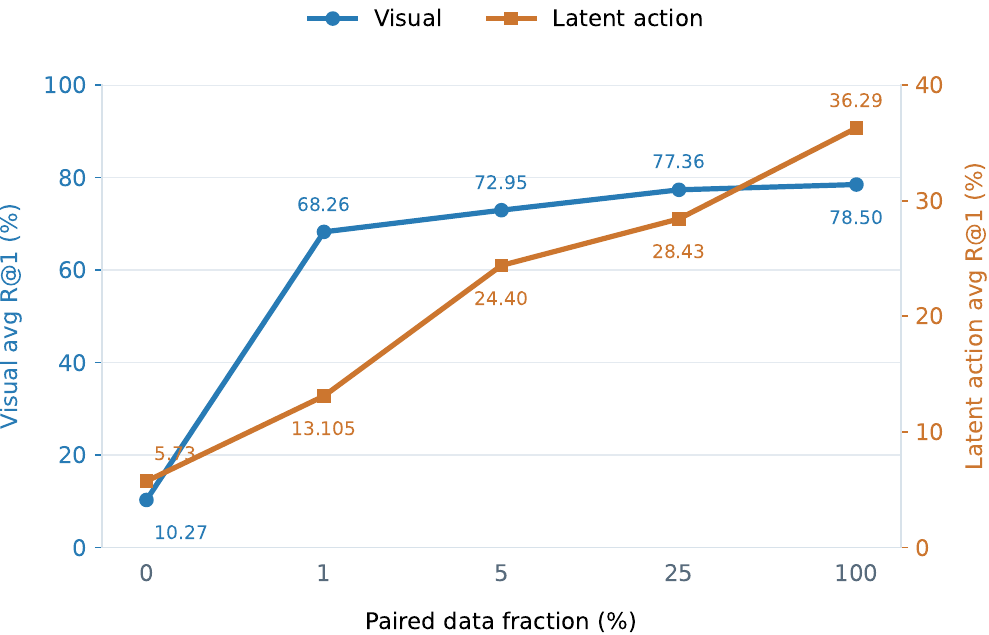}
    \captionof{figure}{Paired-data scaling of visual and latent-action alignment.}
    \label{fig:paired_data_scaling}
\end{minipage}%
\hspace{0.02\linewidth}
\begin{minipage}[t]{0.53\linewidth}
    \vspace{0pt}
    \centering
    \setlength{\tabcolsep}{2pt}
    \renewcommand{\arraystretch}{0.95}
    \small
    \resizebox{\linewidth}{!}{%
        \begin{tabular}{lcccc}
            \toprule
            \textbf{Paired Data} &
            \textbf{Vis. R@1$\uparrow$} &
            \textbf{Lat. R@1$\uparrow$} &
            \textbf{RLBench$\uparrow$} &
            \textbf{CALVIN$\uparrow$} \\
            \midrule
            100\% ($\sim$24M) & \textbf{78.50} & \textbf{36.29} & \textbf{46.59} & \textbf{4.481} \\
            25\% ($\sim$6M)   & 77.36 & 28.43 & 42.10 & 4.354 \\
            5\% ($\sim$1.2M)  & 72.95 & 24.40 & 40.70 & 4.214 \\
            \bottomrule
        \end{tabular}%
    }
    \captionof{table}{%
        % Paired-data scaling of full HARP. Parentheses denote the
        % approximate number of unique paired frames retained;
        % dataset-mixture probabilities, unpaired data, and the
        % optimization budget are fixed. Collecting only 5\% of paired data already yields effective human-robot feature alignment, enabling strong performance on RLBench and CALVIN.
        Paired-data scaling of full HARP across representation alignment and downstream policy evaluations.
        Parentheses denote the approximate number of unique paired frames retained.
        Dataset-mixture probabilities, unpaired data, and the optimization budget are fixed across these nonzero settings.
    }
    \label{tab:paired_scaling}
\end{minipage}%
}

%===============================================================================

\section{Conclusions}

% We presented HARP-VLA, a framework for scalable VLA pretraining from human videos through human-robot aligned visual representations and latent actions.
% Using paired videos as cross-embodiment bridges and unpaired videos as dynamics supervision, HARP improves representation alignment and downstream manipulation performance in both simulation and realworld tasks.
% Future work will scale paired human-robot data to further improve alignment diversity and robustness.

We presented HARP-VLA, a framework which aligns human-robot visual representations and latent actions to improve VLA pretraining from human videos.
By combining paired videos as cross-embodiment bridges, unpaired videos as dynamics supervision, and a source-relative pair-discriminative loss that aligns robot features while preserving pair-level separability, HARP learns an embodiment-aware visual encoder and latent-action model for downstream VLA training.
Experiments on representation alignment, simulation benchmarks, and real-world manipulation show that paired bridge demonstrations, together with unpaired video dynamics supervision, provide an effective route toward VLA pretraining from human videos.

%===============================================================================

\section{Limitations}
% HARP-VLA uses task-paired human-robot videos and auxiliary cues as cross-embodiment bridges, so its performance may depend on the diversity of the bridge data and the robustness of temporal alignment and shared cues.
% Our current evaluation focuses on tabletop manipulation with a single robot platform and a limited set of human-video sources.
% Future work will scale to larger and more diverse human video collections, improve robustness to noisy auxiliary cues, and validate HARP-VLA across more embodiments, longer-horizon tasks and bimanual manipulation.
% 【Update to Camera-Ready】

HARP-VLA requires task-paired human-robot demonstrations and auxiliary cues. Although performance remains strong with fewer unique pairs, collecting paired data remains costly. 
Trajectory-based alignment assumes approximately matched viewpoints and may degrade under tracking errors or large viewpoint/FoV shifts. 
Our implementation is validated with a Prismatic/OpenVLA-OFT-style backbone, while broader architectural generality remains untested. 
Future work will improve cue robustness and evaluate more diverse human videos, embodiments, long-horizon tasks, and bimanual manipulation.

% HARP-VLA relies on task-paired human-robot demonstrations and auxiliary cues as cross-embodiment bridges. 
% Although our scaling study shows strong performance with substantially fewer unique paired demonstrations, the absolute cost of collecting paired data remains non-negligible.
% The trajectory-based alignment currently assumes approximately compatible viewpoints and may degrade under severe tracking errors or large viewpoint/FoV shifts.
% % Our current real-world evaluation is further limited to tabletop manipulation on a single robot platform.
% The current implementation is validated with a Prismatic/OpenVLA-OFT-style backbone, and broader architectural generality remains to be experimentally evaluated.
% Future work will scale to larger and more diverse human video collections, improve robustness to noisy auxiliary cues, and validate HARP-VLA across more embodiments, longer-horizon tasks and bimanual manipulation.

\acknowledgments{We thank the anonymous reviewers for their constructive feedback, which helped improve the clarity and experimental evaluation of this work. We also thank Robotera for providing access to the robotic hardware used in our real-world experiments.}

\bibliography{references}  % .bib

\clearpage
\appendix
\section{Appendix}
\renewcommand{\thetable}{A\arabic{table}}
\setcounter{table}{0}
\renewcommand{\thefigure}{A\arabic{figure}}
\setcounter{figure}{0}

\subsection{Data Process}
\label{app:data}
\input{Appendix/data_process}

\subsection{Loss Details}
\label{app:loss_details}
\input{Appendix/loss_detail}

\subsection{Model Architecture}
\label{app:model}

\textbf{Vision encoder with robot-only adapter.}
We follow Prismatic-7B~\cite{karamcheti2024prismatic} and use a fused vision encoder including DINOv2~\cite{oquab2023dinov2} and SigLIP~ \cite{zhai2023sigmoid} to obtain vision features. By complementing the features of different visual backbones, the generalization and fine-grainedness of visual representations can be improved.

For robot-only adapter, we append a 2-layer MLP after every attention and FFN block in each layer of DINOv2 and SigLIP.
Given the current feature as input, the MLP predicts a residual term that is added back to the original feature. The first MLP layer is Gaussian-initialized, while the second layer is zero-initialized, ensuring the output features are identical to the original features at initialization.

\begin{table}[h]
\centering
\caption{LAM configuration used in Stage-1 training.}
\label{tab:lam_config}
\begin{tabular}{ll}
\toprule
Component & Configuration \\
\midrule
Visual backbone & DINOv2 + SigLIP fused patch tokens \\
Input image size & \(224 \times 224\) \\
Visual tokens per frame & \(16 \times 16 = 256\) \\
Fused token dimension & \(2176\) \\
Transition horizon & randomly sampled \\
Latent-action tokens per transition & \(N_q=4\) \\
VQ codebook size & \(K=16\) \\
Latent dimension & \(d_q=128\) \\
VQ commitment weight & \(\beta=0.25\) \\
Text encoder & frozen T5 encoder \\
Auxiliary tokens & 1 keypoint token, 1 end-effector token \\
\bottomrule
\end{tabular}
\end{table}

\textbf{Latent action model architecture.} 
Table~\ref{tab:lam_config} summarizes the HARP-LAM configuration used in Stage-1 training.
We follow UniVLA~\cite{Bu2025UniVLALT} to construct the latent action model. 
Given an input video clip and a task instruction, the model first converts each frame into visual patch tokens and encodes the language with a text encoder. These visual and language tokens are then fed into a spatiotemporal transformer together with several learned latent query tokens. Among them, the action queries are responsible for capturing the underlying dynamics of the sequence, while additional queries are used to model structured information such as object keypoints and end-effector states.

The core idea of the architecture is to compress action-relevant temporal information into a small set of quantized latent action codes through a vector-quantization bottleneck. These discrete latent codes serve as a high-level abstraction of the motion or behavior needed to explain the video sequence under the given instruction. A decoder then combines the latent action codes with visual context to reconstruct future visual features, while auxiliary heads predict keypoints and end-effector positions to provide extra structural supervision.

\textbf{VLA architecture.} 
We follow OpenVLA-OFT~\cite{kim2025fine} to construct the generalist robot policy backbone, which is built upon Prismatic-7B~\cite{karamcheti2024prismatic}. The architecture contains the same fused vision encoder including SigLIP~\cite{zhai2023sigmoid} and DINOv2~\cite{oquab2023dinov2} in latent action model, a projector aligning fused visual and language embedding space, and a LLaMA-2~\cite{touvron2023llama} LLM backbone. An extra action head is employed in finetuning stages, following OpenVLA-OFT.

Building upon OpenVLA-OFT, we incorporate the robot-only adapter from HARP-LAM into the vision encoder of PrismaticVLM using the same design. Since the fused vision encoder in PrismaticVLM is frozen during web-scale pretraining, the copied weights preserve the same representation alignment properties after initialization.

Moreover, web-scale VLM pretraining is dominated by human-centric data rather than robot data, yielding more semantically robust human representations. Under the HARP framework, the human representation space remains fixed while robot representations are aligned toward the human domain. Consequently, the original alignment between the VLM vision encoder and the LLM is largely preserved, making such initialization well-suited for subsequent VLA training.

\subsection{Human-Robot Representation and Latent-Action Alignment: Experiment Details}
\label{app:align_eval}

\input{Appendix/align_eval}

\subsection{Component Ablation and Paired-Data Scaling Details}
\label{app:ablation_scale}
\input{Appendix/ablation_scale}

\subsection{Robot Policy Learning with Aligned Visual Representations: Experiment Details}
\label{app:rlbench_eval}

\input{Appendix/rlbench_eval}

% \subsection{VLA Policy Evaluation Protocol}
% \label{app:vla_eval}

% For simulation, we evaluate on CALVIN ABC\(\rightarrow\)D following the standard benchmark protocols. 
% For CALVIN, Task1--Task5 report the percentage of trials that complete at least the first \(k\) subtasks in a five-instruction sequence, and Avg. Len. denotes the average number of completed subtasks.

% For real-world evaluation, we evaluate four tabletop manipulation tasks on the Robotera XHand platform. 
% Each method is tested for 60 trials per task under the same task initialization and success criteria, and we report success rates. 
% All HARP-VLA variants use the same downstream real-action demonstrations, action space, action chunk length, and finetuning schedule; they differ only in the pretrained initialization and whether the vision encoder is frozen during VLA pretraining.

% For baselines, we use reproduced or official results under the corresponding benchmark protocols. 
% When reproducing baselines, we use the same downstream demonstrations and evaluation settings as HARP-VLA.

\subsection{VLA Baselines and Real-world Task Details}
\label{app:exp}

% \textbf{Evaluation on Libero Benchmark.} 
% The LIBERO benchmark \cite{liu2023libero} consists of four task suites designed for lifelong robotic manipulation: spatial, object, goal, and long, evaluating capabilities inlcuding spatial reasoning, object generalization, goal-conditioned behavior, and long-horizon multi-step manipulation. Each suite contains 10 tasks with 50 human teleoperated demonstrations per task. We choose the most challenging Libero-long setting as our evaluation task.

\textbf{Evaluation on CALVIN Benchmark.}
The CALVIN benchmark \cite{mees2022calvin} is a language-conditioned long-horizon robotic manipulation benchmark designed to evaluate multi-task and sequential task execution. It consists of a tabletop manipulation environment with diverse objects and scenes, where agents must complete sequences of instructions conditioned on visual observations and natural language commands. Following prior work, we evaluate policies on the most challenging CALVIN ABC$\rightarrow$D setting, where robots are trained with standard datasets collected from environments ABC and tested in the unseen environment D.
For CALVIN, Task1--Task5 report the percentage of trials that complete at least the first \(k\) subtasks in a five-instruction sequence, and Avg. Len. denotes the average number of completed subtasks.

\begin{figure}[t]
  \centering  
  \includegraphics[scale=1.0,width=0.99\linewidth]{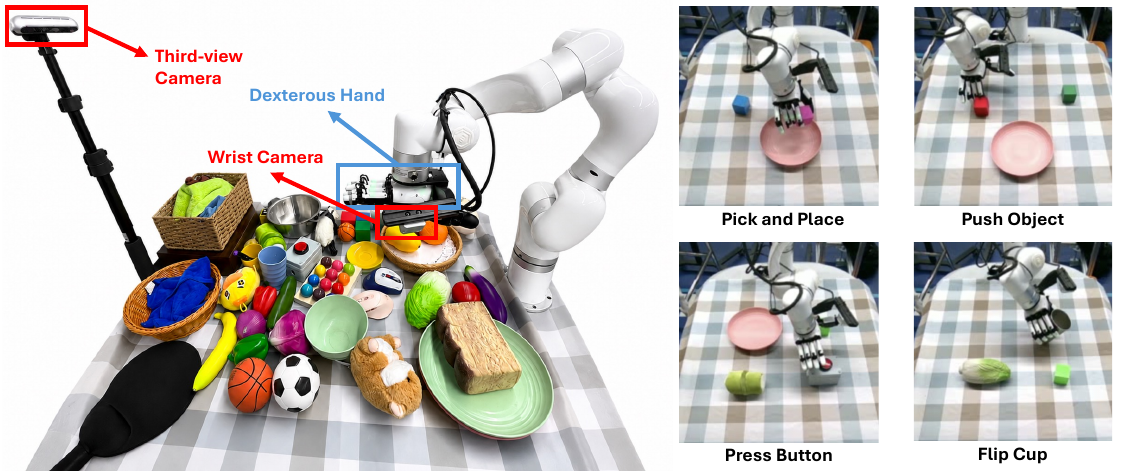} 
  \caption{Real-world platform setup and task overview}
  \label{fig:real_platform}
\end{figure}

\textbf{Evaluation on Real-world Settings.}
% Our realworld experiments are conducted on a robotic platform consisting of an Xarm7 manipulator equipped with a Robotera XHand, resulting in 18 DoF (6 for the Xarm7's end-effector and 12 for the XHand) in total. 
Our policy uses an 18-dimensional action space, consisting of 6D end-effector control for the XArm7 and 12D joint control for the XHand.
The policy takes RGB observations from a third-view camera and a wrist-mounted camera as visual inputs, as well as language instruction, as shown in Fig.~\ref{fig:real_platform}.

To evaluate real-world performance, we choose a set of comprehensive manipulation tasks, including: pick and place, push object, press button, and flip cup.
\begin{itemize}
    \item Pick and Place: the robot was asked to place a specified item into a plate.
    \item Push Object: the robot was asked to push a designated item forward.
    \item Press Button: the robot was asked to press a button.
    \item Flip Cup: the robot was asked to flip the cup upside down and stand it upright.
\end{itemize}
These tasks test HARP-VLA's language grounding, spatial understanding, and dexterity capabilities. 
Each method is tested for 60 trials per task under the same task initialization and success criteria, and we report success rates. 
All HARP-VLA variants use the same downstream real-action demonstrations, action space, action chunk length, and finetuning schedule; they differ only in the pretrained initialization and whether the vision encoder is frozen during VLA pretraining.

\textbf{Baselines.} We selected 5 representative baselines as below, among which UniVLA is similar to our training method, while OpenVLA-OFT is similar to our architecture.
The difference between OpenVLA-OFT and ours is that we freeze the human-robot aligned vision encoder and pre-train only on video data, while OpenVLA-OFT is pre-trained on labeled robot data.
For baselines’ results, we use reproduced or official results under the corresponding benchmark protocols. 
When reproducing baselines, we use the same downstream demonstrations and evaluation settings as HARP-VLA.
\begin{itemize}
    \item \textbf{OpenVLA} \cite{Kim2024OpenVLAAO} is an open-source VLA that learns manipulation skills from robotic datasets by predicting discrete action tokens jointly conditioning on visual observations and natural language instructions.
    \item \textbf{UniVLA} \cite{Bu2025UniVLALT} learns transferable task-centric latent actions from large-scale videos to enable cross-embodiment pre-training for VLAs.
    \item \textbf{OpenVLA-OFT} \cite{kim2025fine} proposes an efficient fine-tuning strategy that improves the inference speed and performance of OpenVLA \cite{Kim2024OpenVLAAO}.
    \item $\boldsymbol{\pi_0}$ \cite{black2026pi0visionlanguageactionflowmodel} is a generalist VLA that generates continuous robot actions via flow-matching.
    \item $\boldsymbol{\pi_{0.5}}$ \cite{intelligence2025pi05visionlanguageactionmodelopenworld} improves $\pi_0$ with enhanced training and scaling to better support open-world robotic manipulation.
\end{itemize}
Following the settings of the original paper itself, in our stage-3 experimental settings, OpenVLA and UniVLA use third-view image as visual input, and OpenVLA-OFT, $\pi_0$, $\pi_{0.5}$ use third-view and wrist view images, which is consistent with HARP.
Except for OpenVLA, which directly outputs the current action, all other baselines output an action chunk of length 10, which is the same as HARP settings.

For the CALVIN experiment, we finetuned $\pi_0$, $\pi_{0.5}$, and OpenVLA-OFT ourselves, and reported the OpenVLA and UniVLA results using publicly available reported results. For the real-world experiment, we finetuned all of them.

%===============================================================================

\clearpage
% The acknowledgments are automatically included only in the final and preprint versions of the paper.

%===============================================================================

% \bibliography{references}  % .bib

\end{document}

%% file: Intro/intro.tex
% 压缩版本idea by GPT
% Introduction 必须保留
% robot demos expensive, human videos abundant；
% action execution gap + visual representation gap；
% LAM mitigates action gap；
% visual gap can make latent actions domain-dependent；
% HARP jointly aligns visual states and latent actions；
% paired bridges + unpaired scalable supervision。

% 这些适合放 related work，不必在 introduction 展开：
% annotation-based methods 为什么 sparse；
% image-level substitution 为什么依赖 sim/hardware；
% HR-Align 为什么主要是 semantic-level；
% LAM 具体有哪些代表工作、区别是什么；
% VLA / human videos for robot learning 的更多背景。

% 可以后移到 Methods, 这些适合 methods，不必在 intro 解释：
% robot-only adapter；
% frozen teacher / preserving loss；
% pair-discriminative alignment；
% keypoint / end-effector auxiliary loss；
% paired/unpaired objective 的具体构成；
% row/col assignment 或 direction loss 的细节。

\begin{figure}[t]
    \centering  
    \includegraphics[width=0.85\linewidth]{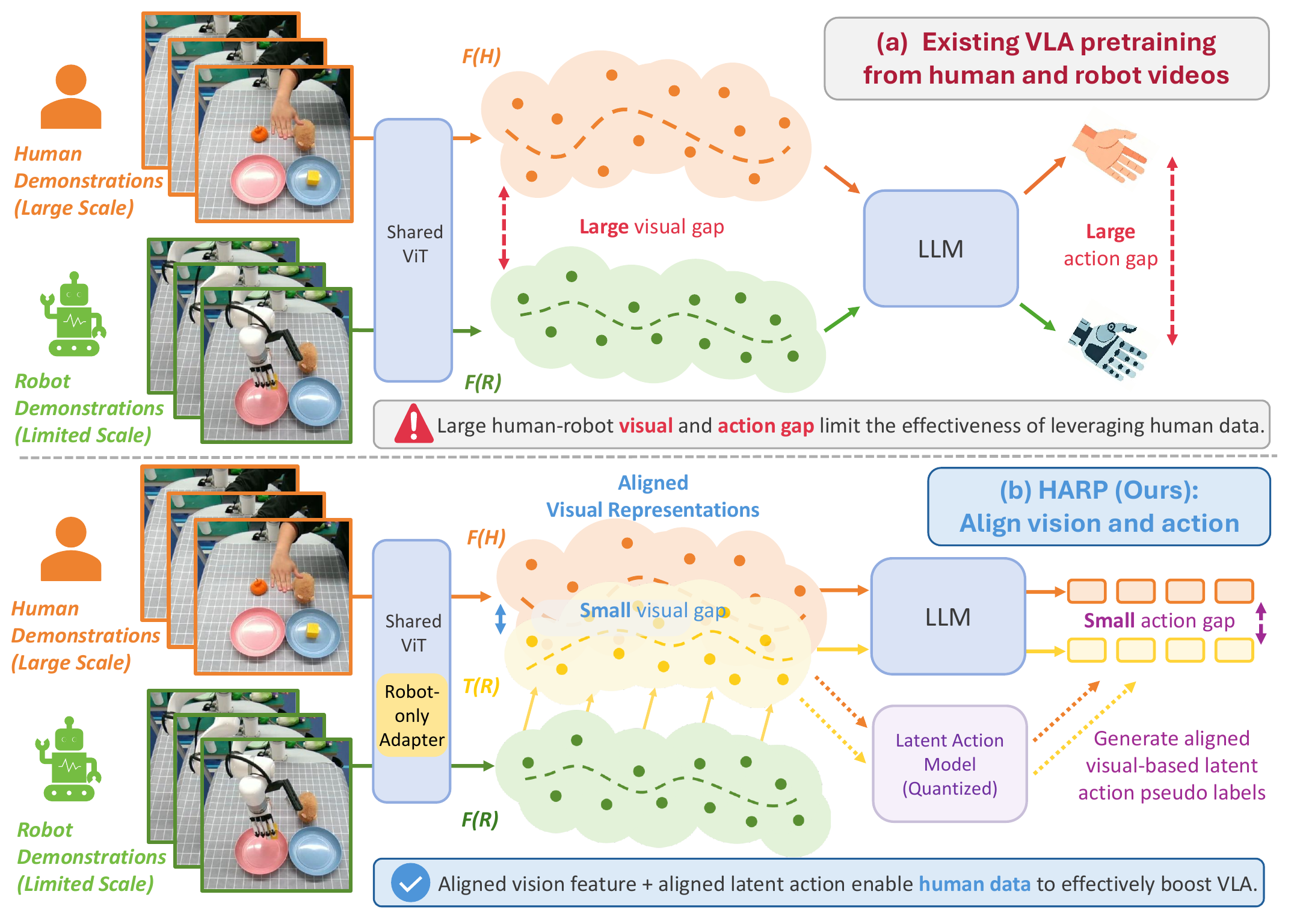} 
\vspace{-0.25em}
\caption{\textbf{Motivation and overview of HARP.} 
\textbf{Top:} Existing VLA pretraining encodes human and robot demonstrations into separated visual representations and suffers from a large action gap, limiting human data use. 
\textbf{Bottom:} HARP jointly aligns visual representations and latent actions using limited paired human-robot demonstrations and scalable unpaired videos, enabling human data to effectively improve VLA pretraining.
}
\vspace{-0.3em}
\label{fig:teaser}
\end{figure}

Learning generalizable robot policies typically requires large-scale teleoperated demonstrations, which are expensive to collect and difficult to scale. 
In contrast, abundant human manipulation videos contain rich skill knowledge, offering a promising source of supervision for robot learning.
However, directly leveraging human videos for robot learning presents two primary challenges. 
The action execution gap makes human motions difficult to translate into executable robot actions, while the visual representation gap causes similar manipulation dynamics to be encoded into separate human and robot feature manifolds,  hindering effective co-training across the two domains, as shown in Fig.~\ref{fig:teaser}(up). 
As a result, policies may learn domain-specific representations rather than reusable skill abstractions, raising a central question: how can we learn a unified policy representation from human and robot videos under cross-domain discrepancies and weak supervision?

Recent advances in Latent Action Models (LAMs)~\cite{Ye2024LatentAP, Bu2025UniVLALT, Kim2025UniSkillIH, Chen2024IGORIR} mitigate the action execution gap by learning latent transition codes from temporally adjacent frames instead of embodiment-specific motor commands. 
% However, latent actions are still grounded in visual observations. 
% If human and robot observations remain visually misaligned, the learned latent actions can become domain-dependent, limiting human-robot co-training and downstream policy learning.
However, because latent actions are grounded in visual observations, misaligned human and robot features can still make them domain-dependent, limiting human-robot co-training and downstream policy learning.
Existing efforts address the visual representation gap through sparse annotations~\cite{Kareer2024EgoMimicSI}, image-level human-to-robot substitution~\cite{li2025mimicdreamer, Xu2025DexUMIUH}, or video-level representation adaptation~\cite{Zhou2024MitigatingTH}. 
However, they do not explicitly couple visual alignment with frame-level manipulation dynamics and latent action learning. 
Therefore, a general and scalable framework is still needed to jointly unify visual states and latent actions across human and robot videos, especially under limited paired supervision and abundant unpaired data.

To address this problem, we propose HARP, a human-robot aligned representation learning framework for robot policy learning from human videos. 
HARP uses limited paired human-robot demonstrations as cross-embodiment bridges and unpaired videos as scalable dynamics supervision. 
It anchors the human-side visual representation to the pretrained encoder, adapts the robot-side encoder with lightweight modules, and jointly learns latent actions over both paired and unpaired videos. 
Together with manipulation-centric auxiliary cues and a source-relative pair-discriminative alignment loss that preserves pair-level discrimination, HARP maps human and robot demonstrations into a unified visual space and latent action space, as shown in Fig.~\ref{fig:teaser}(down).
By reducing cross-domain visual and action discrepancies,  HARP provides a unified interface for policy learning from both human and robot videos, enabling more effective use of human video supervision.
% The learned aligned encoder and latent action model can be integrated into state-of-the-art VLA backbones such as OpenVLA~\cite{Kim2024OpenVLAAO}, where human and robot videos provide vision-language-to-latent-action supervision. 
The learned aligned encoder and latent action model provide a unified interface for VLA pretraining. In this work, we instantiate this interface with a Prismatic/OpenVLA-OFT-style VLA backbone, where human and robot videos provide vision-language-to-latent-action supervision.
A lightweight action head trained on real robot data then translates the learned latent actions into executable robot commands.
The core contributions of this work are threefold:
\begin{enumerate}[leftmargin=1.4em,itemsep=0pt,topsep=1pt,parsep=0pt,partopsep=0pt]
    % \item We propose HARP, a human-robot aligned representation learning framework that enables more effective VLA pretraining from large-scale human videos by jointly training a robot-adapted visual encoder and a latent action model using limited paired demonstrations and scalable unpaired human and robot videos.

    \item We propose HARP, a three-stage framework that jointly learns human-robot aligned visual representations and latent actions for VLA pretraining from large-scale human videos.

    \item We introduce a source-relative pair-discriminative alignment loss tailored to HARP, combined with manipulation-centric auxiliary cues, to align robot representations toward paired human semantics while maintaining pair-level discriminability.

    % \item Extensive experiments in both simulation and real-world robotic settings show that HARP improves human-robot visual and latent-action alignment, and yields consistent gains in downstream VLA-based manipulation policies.
    \item We demonstrate improved visual and latent-action alignment, as well as consistent downstream gains, across representation analysis, simulated benchmarks, and real-world manipulation tasks.

\end{enumerate}

%% file: Methods/data_process.tex
% HARP is a three-stage framework for transferring human demonstration knowledge to robot policy learning through aligned visual representations and latent actions. 
% We first perform \textit{Joint Visual and Latent-Action Alignment} on mixed paired and unpaired manipulation videos, where paired demonstrations provide cross-embodiment correspondences and unpaired videos broaden the coverage of task-relevant dynamics. 
% We then conduct \textit{VLA Pretraining with Aligned Representations}, using the aligned visual encoder and latent-action vocabulary to train a VLA model to predict latent-action tokens from visual observations and language instructions. 
% Finally, in \textit{VLA Finetuning with Real Action}, we fine-tune the model on robot demonstrations and train a lightweight action head to map latent actions to executable continuous robot actions.

HARP is a three-stage framework for transferring human demonstration knowledge to robot policy learning through aligned visual representations and latent actions.
Stage~1 jointly learns a robot-adapted visual encoder and a latent action model (LAM) from paired and unpaired videos; Stage~2 uses the learned encoder and LAM-generated latent-action labels to pretrain a VLA policy; and Stage~3 finetunes the policy on robot demonstrations with a lightweight real-action head. 
Stage 1 is illustrated in Fig.~\ref{fig:stage1}, while Stages 2–3 are shown in Fig.~\ref{fig:pretrain_finetune}.

\subsection{Training Data and Preprocessing}
\textbf{Mixed paired and unpaired demonstrations.} 
Our training data consist of paired and unpaired manipulation videos, 
\(\mathcal{D}=\mathcal{D}_p\cup\mathcal{D}_u\). 
The paired set is defined as 
\(
\mathcal{D}_p=\{(H_i,R_i,l_i)\}_{i=1}^{N_p}, 
\)
where \(H_i\) and \(R_i\) are paired human and robot videos of the same task with instruction \(l_i\). 
The unpaired set is \( \mathcal{D}_u=\{(V_j,l_j)\}_{j=1}^{N_u}, \) where \(V_j\) is a video from embodiment \(e_{V_j}\in\{h,r\}\). % with instruction \(l_j\). 
Paired videos provide cross-embodiment supervision, while unpaired videos enrich task-relevant dynamics.

\textbf{Shared task cues across embodiments.} 
Human and robot demonstrations differ in appearance and morphology, but often share task-level cues such as instructions, object motion, and coarse agent motion.
% These cues motivate our auxiliary objectives, which regularize the learned representations and latent actions toward shared task dynamics rather than embodiment-specific appearance. 
For each video \(X\), we extract auxiliary annotations 
\(
A_X=\{K_X,E_X\}, % A_X=\{K_X,E_X,M_X\}, 
\)
where \(K_X\) denotes 2D object position tracks, \(E_X\) denotes 2D human wrist or robot end-effector trajectory. % and \(M_X\) denotes the corresponding visibility masks. 
Object tracks capture manipulation effects, while wrist/end-effector trajectories provide a coarse proxy for motion intent. 
% These cues motivate our auxiliary objectives, which regularize the learned representations and latent actions toward shared task dynamics rather than embodiment-specific appearance and also serve as weak embodiment-shared supervision, especially for unpaired videos where explicit cross-embodiment positives are unavailable. 
These cues regularize the latent actions to capture motion dynamics rather than embodiment-specific appearance, especially for unpaired videos without cross-embodiment supervision.

\textbf{Temporally aligned paired videos.} 
For paired videos, although similar motion patterns are enforced during data collection, execution speed and subtask duration remain difficult to keep consistent. 
% Thus, we perform dynamic time warping (DTW) over \(E_X\) to obtain temporal correspondence maps, warping human videos onto the robot timeline for stricter progress alignment.
Thus, we perform dynamic time warping (DTW) over \(E_X\) to obtain temporal correspondence maps and sample matched human-robot transitions on the robot timeline.

Finally, processed
\(
\mathcal{D}_p=\{(\hat{H}_i,\hat{R}_i,l_i,A_{\hat{H}_i},A_{\hat{R}_i})\}_{i=1}^{N_p}
\),
where $\hat{\cdot}$ denotes temporally aligned, which is omitted hereafter for simplicity, and
\(
\mathcal{D}_u=\{(V_j,l_j,A_{V_j})\}_{j=1}^{N_u} 
\).
% We therefore perform dynamic time warping (DTW) over annotated motion trajectories to obtain a temporal correspondence map \(\alpha_i\).
% A human transition 
% \((h_i^t,h_i^{t+\Delta})\) 
% is paired with the robot transition 
% \((r_i^{\alpha_i(t)},r_i^{\alpha_i(t+\Delta)})\). 
% For any video \(X\), a sampled transition is written as 
% \( c_X^t=(x_X^t,x_X^{t+\Delta},l_X,A_X^t,A_X^{t+\Delta}). \) 
Paired videos support cross-embodiment prediction and alignment, while unpaired videos support self-prediction and auxiliary supervision.
Details of annotation extraction and DTW alignment are provided in Appendix~\ref{app:data}.

%% file: Methods/ve_lam_joint_training_new.tex
\subsection{Joint Visual and Latent-Action Alignment}
% \subsection{Embodiment-Aligned Latent Action Learning}
\label{sec:lam}

\paragraph{Embodiment-aware visual encoding.}
Given a video \(X\) with embodiment \(e_X\in\{h,r\}\), we encode frames into patch-level visual tokens using an embodiment-aware encoder composed of the frozen original visual encoder \(F\) and robot-adapted encoder \(T_\theta\) (see Appendix~\ref{app:model} for architecture),
% Let \(F\) be the frozen original visual encoder and \(T_\theta\) be the robot-adapted encoder (see Appendix~\ref{app:model} for architecture). We define the unified embodiment-aware encoder as
\[
\Phi_\theta(X,e_X)=
\begin{cases}
F(X), & e_X=h,\\
T_\theta(X), & e_X=r.
\end{cases}
\]
The resulting visual tokens and frozen teacher features are denoted by
\[
Z_X=\Phi_\theta(X,e_X)=\{z_X^t\}_{t=1}^{T_X}, 
\qquad 
Z_{X0}=F(X),
\]
where $T_X$ is the total frame number. This design anchors human videos to the pretrained visual space while adapting robot representations toward the same semantic space.
% This formulation keeps the human representation anchored to the frozen human-pretrained encoder, while allowing the robot representation to be adapted toward the human semantic space.

\paragraph{Latent-action prediction with Self- and Cross-Prediction.}
We instantiate the latent action module as a VQ-VAE-style inverse-and-forward dynamics model, where an encoder infers latent actions from visual transitions, and a decoder predicts future visual representations conditioned on the current observation and the quantized latent action. 
Each transition is represented by \(N_q=4\) discrete latent-action tokens quantized by a VQ codebook.
Here, \(t+\Delta t\) denotes the second frame in the sampled transition; details are provided in Appendix~\ref{app:model}.

Given a transition \((z_X^t,z_X^{t+\Delta t})\) and language instruction \(l_X\), the encoder $E_\theta$ infers a continuous latent action, quantizes it with a codebook \(Q_\theta\), and the decoder $D_\theta$ predicts the target frame's patch-level visual tokens as the future representation, and \(\tilde{z}_X^t\) denotes the decoder conditioning feature.
\[
a_X^t=E_\theta(z_X^t,z_X^{t+\Delta t},l_X), 
\qquad 
q_X^t=Q_\theta(a_X^t), 
\qquad 
\hat{Y}_X^t = D_\theta(\tilde{z}_X^t,q_X^t,l_X),
\]

The decoder target \(Y_X^t\) and condition \(\tilde{z}_X^t\) depend on whether the transition is sampled from paired or unpaired data. For an unpaired video \(V\), HARP uses self-prediction: the model predicts the future representation within the same video
\(
\tilde{z}_V^t=z_V^t, Y_V^t=z_V^{t+\Delta t}.
\)
For paired videos \((H,R)\), HARP performs cross-embodiment prediction.
% Using the temporally aligned human and robot transitions, with aligned indices written as \(t\) for simplicity, 
The latent action is inferred from the source transition, while the decoder is conditioned on the current representation of the target embodiment:
\[
\tilde{z}_H^t=z_R^{t}, \quad 
Y_H^t=z_R^{t+\Delta t};
\qquad
\tilde{z}_R^t=z_H^{t}, \quad 
Y_R^t=z_H^{t+\Delta t}.
\]
% Here the frame indices are selected according to the offline temporal correspondence described in Sec.~\ref{sec:data}. 
This cross-prediction objective forces latent actions extracted from one embodiment to explain future dynamics of the other, thereby coupling latent-action learning with human-robot alignment.

\paragraph{Training objective.}
Stage-1 training combines three objectives:
\[
\mathcal{L}_{\mathrm{stage1}}
=
\mathcal{L}_{\mathrm{lam}}
+
\lambda_{\mathrm{aux}}\mathcal{L}_{\mathrm{aux}}
+
\lambda_{\mathrm{align}}\mathcal{L}_{\mathrm{align}} .
\]
The latent action prediction loss learns discrete latent actions through self- and cross-prediction, the auxiliary loss injects manipulation-centric shared cues, and the alignment loss consists of a source-relative term and a pair-discriminative regularizer to adapt robot representations toward human semantics while maintaining pair-level discrimination.

\textbf{Latent-action prediction.}
Given the decoder prediction \(\hat{Y}_X^t\) and target \(Y_X^t\), we optimize
\[
\mathcal{L}_{\mathrm{lam}}
=\mathbb{E}_{t < T_X, X \sim \mathcal{D}}
\|\hat{Y}_X^t-Y_X^t\|_2^2
+
\mathcal{L}_{\mathrm{vq}},
\]
where \(\mathcal{L}_{\mathrm{vq}}\) is the standard VQ codebook and commitment loss.
This term trains the latent action to explain future visual dynamics under both self-prediction and cross-embodiment prediction.

\textbf{Shared-cue auxiliary loss.}
To encourage task-relevant dynamics, we regularize latent-action learning with object-centric point tracks and wrist/end-effector trajectories. We introduce shared-cue auxiliary tokens in the latent-action encoder to predict these shared cues, and supervise them with
\[
\mathcal{L}_{\mathrm{aux}}
=
\lambda_K\mathcal{L}_K
+
\lambda_E\mathcal{L}_E .
\]
Here, \(\mathcal{L}_K\) and \(\mathcal{L}_E\) are Huber losses over visible object keypoints and wrist/end-effector positions, respectively. These losses bias the latent actions toward object motion and coarse agent motion, which are more shared across embodiments than raw appearance. 

\textbf{Source-relative pair-discriminative alignment loss.}
For paired demonstrations, the adapted robot representation should satisfy two constraints: it should improve over the frozen robot representation with respect to the paired human video, and the paired correspondence should remain discriminative against non-matching pairs.
For paired videos \((H,R)\), let
\(
f^H=\rho(Z_H), 
f^{R0}=\rho(Z_{R0}), 
f^R=\rho(Z_R),
\)
where \(\rho(\cdot)\) denotes video-level pooling.

Using cosine distance \(d(u,v)=1-\cos(u,v)\), the source-relative term requires the adapted robot feature to improve over the frozen robot feature with respect to the paired human feature:
\[
\mathcal{L}_{\mathrm{SR}}
=\mathbb{E}_{(H,R)\sim \mathcal{D}_p}
\left[
m_s
+
d(f^R,f^H)
-
d(f^{R0},f^H)
\right]_+,
\]
where $m_s$ is a fixed triplet margin. 
This source-relative formulation avoids forcing all paired features to collapse to an absolute distance target.

% We further add a pair-discriminative regularizer \(\mathcal{L}_{\mathrm{PD}}\), which uses non-matching pairs in the batch as negatives to preserve cross-pair separability. 
To preserve pair-level discrimination, we further add a pair-discriminative term
% \[
% \mathcal{L}_{\mathrm{PD}}
% =
% \mathbb{E}_i
% \sum_{\alpha\in\{\mathrm{R2H},\mathrm{H2R}\}}
% \lambda_{\alpha}
% [m_t+d(f_i^R,f_i^H)-\bar d_i^{\alpha}]_+ ,
% \]
% where
% \(\bar d_i^{\mathrm{R2H}}=\mathrm{mean}_{j\ne i}d(f_i^R,f_j^H)\) and
% \(\bar d_i^{\mathrm{H2R}}=\mathrm{mean}_{j\ne i}d(f_j^R,f_i^H)\).
\[
\mathcal{L}_{\mathrm{PD}}
=
\mathbb{E}_{(H,R)\sim\mathcal D_p}
\textstyle \sum_{\alpha\in\{\mathrm{R\!\to\!H},\mathrm{H\!\to\!R}\}}
\lambda_{\alpha}
\left[
m_t+d(f^R,f^H)-\bar d^\alpha
\right]_+ ,
\]
where
\(
\bar d^{\mathrm{R\!\to\!H}}
=
\mathbb{E}_{H'\neq H}
d(f^R,f^{H'}),
\bar d^{\mathrm{H\!\to\!R}}
=
\mathbb{E}_{R'\neq R}
d(f^{R'},f^H),
\)
and $m_t$ is a fixed margin.

% \[
% \mathcal{L}_{\mathrm{PD}}
% =
% \mathbb{E}_{(H,R)\sim\mathcal D}
% \Big[
% \lambda_1
% \big(
% m_t+d(f^R,f^H)
% -
% \mathbb{E}_{H^-\neq H} d(f^R,f^{H^-})
% \big)_+
% \]
% \[
% +
% \lambda_2
% \big(
% m_t+d(f^R,f^H)
% -
% \mathbb{E}_{R^-\neq R} d(f^{R^-},f^H)
% \big)_+
% \Big].
% \]

The final alignment loss is
\[
\mathcal{L}_{\mathrm{align}}
=
\mathcal{L}_{\mathrm{SR}}
+
\mathcal{L}_{\mathrm{PD}}.
\]
Detailed formulations of the training objectives are provided in Appendix~\ref{app:loss_details}. We refer to the latent action model trained as HARP-LAM, and the aligned vision encoder as HARP-VE.

%% file: Appendix/data_process.tex
\begin{figure}[t]
  \centering  
  \includegraphics[scale=1.0,width=0.99\linewidth]{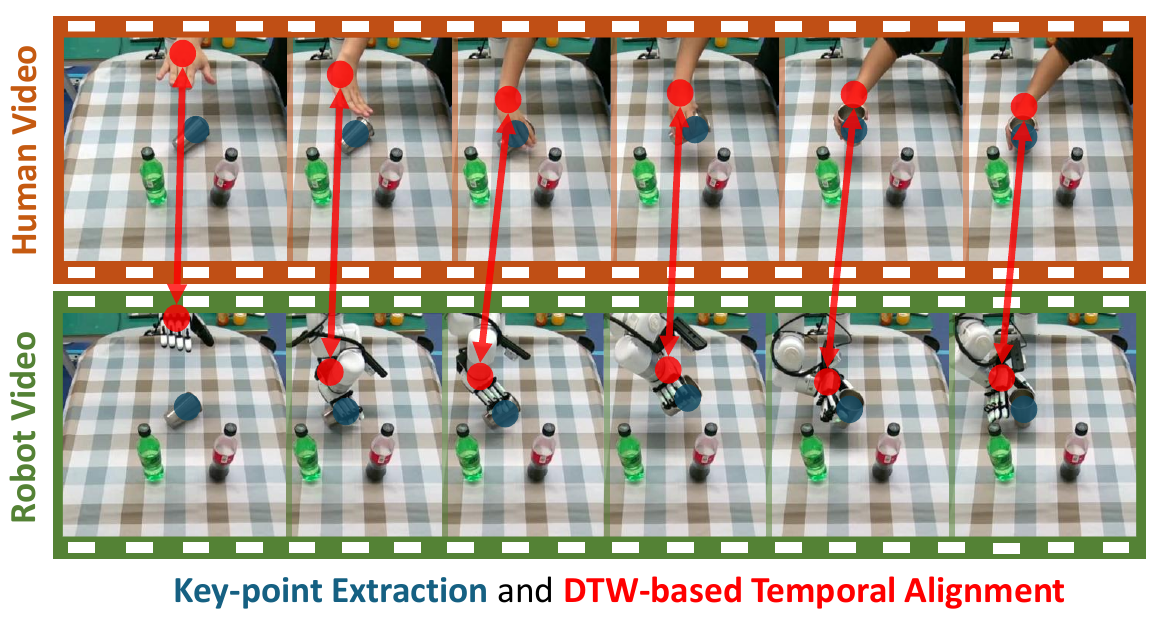} 
  \caption{Paired data curation pipeline. We extract object-keypoint (blue points) for extra supervision, wrist position (red points) for supervision, and L2-distance-based dynamic time warping (red arrows).}
  \label{fig:pair_data}
\end{figure}

The objective of paired training data curation is to establish frame-level correspondence between human demonstration data and robot motion data, in order to provide frame-level visual clues for the following cross-prediction procedure. 
We propose to extract different types of keypoints as extra supervision and perform temporal alignment to better synchronize motion patterns for cross-prediction. As shown in Fig.~\ref{fig:pair_data}.

\paragraph{Keypoint extraction.}
For all human and robot videos, we extract two kinds of keypoints for supervision: object keypoint following object-flow-based methods \cite{Xu2024FlowAT, dharmarajan2025dream2flow} and hand keypoint following bridge-domain-gap methods \cite{Kareer2024EgoMimicSI, zhu2025learninggeneralizablerobotpolicy}.

Regarding object keypoint extraction, our pipeline is explicitly designed to handle severe occlusion, which is largely underexplored in prior work. If no language instruction is available, we first use Qwen3-VL-8B-Instruct \cite{qwen3technicalreport} to generate a description of the manipulated object, then apply GroundingDINO \cite{liu2024grounding} to localize its bounding box at the first occurrence, extract keypoints from this initial frame, and track them with TAPIR \cite{doersch2023tapir}. Although this design appears complex, it is necessary because existing pipelines typically fail under full occlusion and rarely address it explicitly. TAPIR is particularly suitable as it leverages historical context to infer current positions and provides occlusion and visibility scores, which enable us to attenuate unreliable predictions during occluded frames and thus improve overall robustness. 

Towards hand keypoint extraction in human videos, we employ WiLoR \cite{potamias2025wilor}, a monocular 3D hand reconstruction framework that regresses MANO \cite{romero2022embodied} parameters from RGB input, from which we directly obtain the wrist’s 2D coordinates on the image plane. As for hand keypoint extraction in robot videos, we leverage the camera extrinsics together with the recorded wrist pose in 3D space to project it onto the image plane, thereby computing the corresponding 2D coordinates in each frame through standard perspective projection.

% 得提一嘴采集paired data成本很低，也不用熟练员工
\paragraph{Temporal alignment.}
For human–robot paired videos, although similar motion patterns are enforced during data collection, the execution speed and duration of subtask actions remain difficult to keep strictly consistent. 

To address this temporal discrepancy, we use the Euclidean distance between the previously extracted hand keypoints as a similarity metric to characterize the correspondence between human and robot videos. We then apply Dynamic Time Warping (DTW) \cite{yadav2018dynamic} to achieve strict frame-wise temporal alignment.
Specifically, we take the more uniformly executed robot video as the temporal reference and resample the human video by filtering or duplicating frames according to the optimal DTW matching path. 
In this way, we obtain fully synchronized human-robot paired videos in terms of both duration and execution speed, reducing the need for precise temporal alignment during human paired-data collection and allowing the data acquisition process to focus primarily on matching motion patterns.

\paragraph{Training data summary.}
Table~\ref{tab:data_summary} summarizes the data and supervision used in each training stage. 
Scale denotes the number of frames used after preprocessing and filtering. 
For RH20T, due to heterogeneous camera viewpoints and variable pair quality, we use a selected subset rather than the full raw paired-video pool.
For Ours-Real, only video observations are used in Stage~1/2, while real action labels are used only in Stage~3. Evaluation trials are held out from finetuning demonstrations.

\begin{table}[h]
\centering
\caption{
Dataset-level data usage in HARP-VLA.
S1 denotes HARP-LAM/HARP-VE training, S2 denotes VLA latent-action pretraining, and S3 denotes downstream real-action finetuning.
}
\label{tab:data_summary}
\resizebox{\linewidth}{!}{
\begin{tabular}{llllcccl}
\toprule
Dataset & Embodiment & Data type & Scale used & S1 & S2 & S3 \\
\midrule
HOI4D & Human & Unpaired video &  8.9M frames & Yes & Yes & No \\
OpenEgo & Human & Unpaired video &  36.4M frames & Yes & Yes & No \\
Bridge-V2 & Robot & Unpaired video &  8.6M frames & Yes & Yes & No \\
Ours-Unpaired & DexHand & Unpaired video &  3.6M frames & Yes & Yes & No \\
RH20T & Human-Robot & Paired video &  8.16M frames & Yes & Yes & No \\
Human2Robot & Human-Robot & Paired video &  9.9M frames & Yes & Yes & No  \\
Ours-Paired & Human-DexHand & Paired video & 5.7M frames & Yes & Yes & No  \\
CALVIN & Robot / Sim. & Real-action demos & 1.1M frames & No & No & Yes  \\
Ours-Real & DexHand / Real & Real-action demos & 0.5M frames & Yes & Yes & Yes \\
\bottomrule
\end{tabular}
}
\end{table}

%% file: Appendix/loss_detail.tex
This section provides the detailed formulations of the losses used in Stage~1. 
We denote a sampled transition from video \(X\) as \((z_X^t,z_X^{t+\Delta t})\), with language instruction \(l_X\), where \(t+\Delta t\) denotes the second frame in the sampled transition.
The latent-action encoder produces continuous latent-action vectors \(a_X^t\), which are quantized into \(q_X^t\).
The decoder prediction is denoted as \(\hat{Y}_X^t\), and the corresponding target is \(Y_X^t\), as defined in Sec.~\ref{sec:lam}.

Given a transition, the decoder predicts
\[
a_X^t=E_\theta(z_X^t,z_X^{t+\Delta t},l_X),
\qquad
q_X^t=Q_\theta(a_X^t),
\qquad
\hat{Y}_X^t=D_\theta(\tilde z_X^t,q_X^t,l_X),
\]
where \(\tilde z_X^t\) is the decoder conditioning feature.
For an unpaired video \(V\), HARP uses self-prediction:
\[
\tilde z_V^t=z_V^t,
\qquad
Y_V^t=z_V^{t+\Delta t}.
\]
For a paired sample \((H_i,R_i)\), HARP uses cross-embodiment prediction:
\[
\tilde z_{H_i}^t=z_{R_i}^t,
\quad
Y_{H_i}^t=z_{R_i}^{t+\Delta t};
\qquad
\tilde z_{R_i}^t=z_{H_i}^t,
\quad
Y_{R_i}^t=z_{H_i}^{t+\Delta t}.
\]

\paragraph{Latent action prediction loss.}
The latent action prediction loss contains a future-representation prediction term and a standard VQ loss. 
For a transition from video \(X\), we define
\[
\ell_{\mathrm{lam}}(X,t)
=
\left\|
\hat{Y}_X^t-Y_X^t
\right\|_2^2
+
\ell_{\mathrm{vq}}(X,t).
\]
The VQ loss is
\[
\ell_{\mathrm{vq}}(X,t)
=
\left\|
\mathrm{sg}[a_X^t]-q_X^t
\right\|_2^2
+
\beta
\left\|
a_X^t-\mathrm{sg}[q_X^t]
\right\|_2^2 ,
\]
where \(\mathrm{sg}[\cdot]\) denotes the stop-gradient operator and \(\beta\) is the commitment weight. 
% For unpaired videos, \(Y_X^t\) is the future representation from the same video. 
% For paired videos, \(Y_X^t\) is taken from the temporally aligned counterpart embodiment, enabling cross-embodiment prediction.

Given a set of sampled transitions \(\mathcal{B}\), the batch-level latent action prediction loss is
\[
\mathcal{L}_{\mathrm{lam}}
=
\frac{1}{|\mathcal{B}|}
\sum_{(X,t)\in\mathcal{B}}
\ell_{\mathrm{lam}}(X,t).
\]

\paragraph{Masked shared-cue auxiliary loss.}
To inject manipulation-centric cues, we introduce dedicated auxiliary tokens in the latent-action encoder. 
Let \(u_{X,K}^{\tau}\) and \(u_{X,E}^{\tau}\) denote the auxiliary token embeddings for object-centric keypoints and wrist/end-effector prediction at frame \(\tau\). 
Two lightweight prediction heads produce
\[
\hat{K}_{X}^{\tau}=G_K(u_{X,K}^{\tau}),
\qquad
\hat{E}_{X}^{\tau}=G_E(u_{X,E}^{\tau}),
\]
where \(\hat{K}_{X}^{\tau}\in\mathbb{R}^{N_K\times 2}\) denotes predicted object-centric point locations, and \(\hat{E}_{X}^{\tau}\in\mathbb{R}^{2}\) denotes the predicted wrist position for human videos or end-effector position for robot videos.

Let \(K_X^\tau\in\mathbb{R}^{N_K\times 2}\) and \(E_X^\tau\in\mathbb{R}^{2}\) be the corresponding annotations. 
We denote their visibility masks as \(M_{X,K}^{\tau,k}\in\{0,1\}\) and \(M_{X,E}^{\tau}\in\{0,1\}\). 
Using the Huber loss \(\mathcal{H}(\cdot,\cdot)\), the masked keypoint loss is
\[
\ell_K(X)
=
\frac{
\sum_{\tau}
\sum_{k=1}^{N_K}
M_{X,K}^{\tau,k}
\,
\mathcal{H}
\left(
\hat{K}_{X,k}^{\tau},
K_{X,k}^{\tau}
\right)
}{
\sum_{\tau}
\sum_{k=1}^{N_K}
M_{X,K}^{\tau,k}
+
\epsilon
},
\]
and the masked wrist/end-effector loss is
\[
\ell_E(X)
=
\frac{
\sum_{\tau}
M_{X,E}^{\tau}
\,
\mathcal{H}
\left(
\hat{E}_{X}^{\tau},
E_X^{\tau}
\right)
}{
\sum_{\tau}
M_{X,E}^{\tau}
+
\epsilon
}.
\]
The auxiliary loss is
\[
\ell_{\mathrm{aux}}(X)
=
\lambda_K\ell_K(X)
+
\lambda_E\ell_E(X),
\]
and the batch-level objective is
\[
\mathcal{L}_{\mathrm{aux}}
=
\frac{1}{|\mathcal{B}|}
\sum_{X\in\mathcal{B}}
\ell_{\mathrm{aux}}(X).
\]
Here, \(\epsilon\) is a small constant for numerical stability.

\paragraph{Source-relative pair-discriminative alignment loss.}
The alignment loss is applied to paired human-robot demonstrations.
For a paired batch \(\mathcal{B}_p=\{(H_i,R_i)\}_{i=1}^{B}\), we define
\[
f_i^H=\rho(Z_{H_i})=\rho(Z_{H_i0}),
\qquad
f_i^{R0}=\rho(Z_{R_i0}),
\qquad
f_i^R=\rho(Z_{R_i}),
\]
where \(f_i^H\) is the frozen human representation, \(f_i^{R0}\) is the frozen robot representation, and \(f_i^R\) is the adapted robot representation.
All features are \(\ell_2\)-normalized before computing cosine distances.
We use
\[
d(u,v)=1-\cos(u,v),
\qquad
d_i^+=d(f_i^R,f_i^H).
\]

\emph{Source-relative term.}
The source-relative term requires the adapted robot feature to improve over the frozen robot feature with respect to the paired human feature:
\[
\ell_{\mathrm{SR}}(i)
=
\left[
m_s
+
d_i^+
-
d(f_i^{R0},f_i^H)
\right]_+ ,
\]
where \([x]_+=\max(x,0)\), and \(m_s\) is the source-relative margin.

\emph{Pair-discriminative term.}
To maintain pair-level discrimination, we use non-matching pairs in the batch as negatives.
For \(B>1\), the mean negative distances in the two matching directions are
\[
\bar{d}^{R\rightarrow H}(i)
=
\frac{1}{B-1}
\sum_{j\neq i}
d(f_i^R,f_j^H),
\qquad
\bar{d}^{H\rightarrow R}(i)
=
\frac{1}{B-1}
\sum_{j\neq i}
d(f_j^R,f_i^H).
\]
The corresponding pair-discriminative losses are
\[
\ell_{\alpha}(i)
=
\left[
m_t
+
d_i^+
-
\bar{d}^{\alpha}(i)
\right]_+,
\qquad
\alpha\in\{R\rightarrow H,\ H\rightarrow R\},
\]
where \(m_t\) is the pair-discrimination margin.
We summarize the two directions as
\[
\ell_{\mathrm{PD}}(i)
=
\lambda_{R\rightarrow H}\ell_{R\rightarrow H}(i)
+
\lambda_{H\rightarrow R}\ell_{H\rightarrow R}(i).
\]
The alignment loss for pair \(i\) is
\[
\ell_{\mathrm{align}}(i)
=
\ell_{\mathrm{SR}}(i)
+
\ell_{\mathrm{PD}}(i).
\]
The batch-level alignment loss is
\[
\mathcal{L}_{\mathrm{align}}
=
\frac{1}{|\mathcal{B}_p|}
\sum_{i=1}^{|\mathcal{B}_p|}
\ell_{\mathrm{align}}(i).
\]
When \(B=1\), the pair-discriminative term is omitted and only the source-relative term is used.

The full Stage-1 objective is
\[
\mathcal{L}_{\mathrm{stage1}}
=
\mathcal{L}_{\mathrm{lam}}
+
\lambda_{\mathrm{aux}}\mathcal{L}_{\mathrm{aux}}
+
\lambda_{\mathrm{align}}\mathcal{L}_{\mathrm{align}}.
\]

%% file: Appendix/align_eval.tex
We evaluate representation and latent-action alignment on 93 held-out paired human-robot demonstrations that are not used for training.
For all methods, we use the same frame sampling strategy, video-level pooling function \(\rho(\cdot)\), feature normalization, and cosine distance/similarity metrics.
Unless otherwise specified, we use the fused DINOv2+SigLIP feature branch.
Patch tokens are first averaged spatially within each frame and then temporally averaged over sampled frames to obtain a video-level embedding, followed by \(\ell_2\)-normalization.

\paragraph{UMAP visualization.}
For visual-representation UMAP, we compare the original feature space \(F(H)\) versus \(F(R)\) with the adapted feature space \(F(H)\) versus \(T_\theta(R)\).
Human and robot embeddings from the same held-out paired set are projected with UMAP using cosine distance.
Circles denote human videos and triangles denote robot videos; gray points show sampled pairs and colored points highlight a small subset of pairs for readability.
For latent-action UMAP, we extract the quantized latent-action embeddings from HARP-LAM.
The baseline panel uses latent actions computed from the original visual encoder for both human and robot videos, while the HARP-LAM panel uses the frozen human branch and the adapted robot branch.
The latent-action tokens are flattened and \(\ell_2\)-normalized before UMAP projection.

\paragraph{Paired cosine distance.}
For paired cosine distance, we compute
\[
d_i^{\mathrm{orig}}
=
1-\cos(\rho(F(H_i)),\rho(F(R_i)))
\]
for the unadapted encoder, and
\[
d_i^{\mathrm{adapt}}
=
1-\cos(\rho(F(H_i)),\rho(T_\theta(R_i)))
\]
for adapted encoders.
Fig.~\ref{fig:paired_distance} visualizes the distribution of pair-level distances over the 93 held-out pairs using box plots, where lower values indicate smaller human-robot representation gaps.

\paragraph{Cross-embodiment retrieval.}
For bidirectional retrieval, H2R uses each human video as a query and ranks all 93 robot videos by cosine similarity, while R2H uses each robot video as a query and ranks all 93 human videos.
The matched demonstration is the diagonal positive pair in the similarity matrix.
We report Recall@1 in the main paper and provide additional retrieval metrics, including mean reciprocal rank (MRR), in Table~\ref{tab:retrieval_appendix}.

\begin{table}[h]
\centering
\caption{
Supplementary bidirectional cross-embodiment retrieval metrics on held-out paired demonstrations.
}
\label{tab:retrieval_appendix}
\begin{tabular}{lcccccc}
\toprule
Method & H2R R@1 & H2R MRR & R2H R@1 & R2H MRR & Avg. R@1 & Avg. MRR \\
\midrule
Unadapted & 44.09 & 0.5936 & 43.01 & 0.5882 & 43.55 & 0.5909 \\
HR & 45.16 & 0.6022 & 45.16 & 0.6269 & 45.16 & 0.6145 \\
HARP-HR & 46.24 & 0.6286 & 60.22 & 0.7398 & 53.23 & 0.6842 \\
HARP-L2 & 70.97 & 0.8310 & 52.69 & 0.6742 & 61.83 & 0.7526 \\
HARP-SR & 84.95 & 0.9091 & 64.52 & 0.7680 & 74.74 & 0.8385 \\
HARP-SRPD & \textbf{87.10} & \textbf{0.9283} & \textbf{69.89} & \textbf{0.8010} & \textbf{78.50} & \textbf{0.8647} \\
\bottomrule
\end{tabular}
\end{table}

% 【Update for Camera-ready】
\paragraph{Cross-embodiment latent-action correspondence.} 
For latent-action retrieval, we extract quantized latent-action embeddings for temporally aligned human and robot transitions from the same held-out paired pool. The latent-action tokens are flattened and $\ell_2$-normalized. H2R retrieval uses each human transition as a query to rank robot transitions by cosine similarity, while R2H is defined symmetrically, with the temporally corresponding transition treated as the positive. We report the average H2R/R2H Recall@1 as Lat. R@1.

We additionally measure discrete code agreement (CA). For each temporally corresponding human-robot transition, CA is the fraction of the $N_q$ latent-action slots assigned the same VQ code. To discount accidental agreement caused by global code usage, we randomly shuffle robot samples across human samples and recompute the agreement. We define
$$ \mathrm{CA\text{-}gap} = \mathrm{CA}_{\mathrm{paired}} - \mathrm{CA}_{\mathrm{shuffled}}. $$
A larger CA-gap indicates stronger correspondence-specific agreement of discrete latent actions across embodiments.

%% file: Appendix/ablation_scale.tex
\paragraph{Component ablations.} 
We perform controlled Stage-1 ablations while keeping all other training and evaluation settings fixed. For paired samples, Self-only uses only self-prediction, Cross-only uses only cross-prediction, and Full HARP combines both; unpaired samples always use self-prediction. Paired-only removes unpaired training videos, while w/o Aux removes object-keypoint and wrist/end-effector supervision. Across these variants, codebook perplexity remains 10.0--14.2 out of a maximum of 16, indicating substantial codebook utilization rather than collapse.

\paragraph{Paired-data scaling.} 

For nonzero paired-data fractions, we uniformly subsample the unique examples in each paired human-robot dataset to 25\%, 5\%, or 1\% of the full paired pool, corresponding to approximately 6M, 1.2M, and 0.24M retained paired frames, respectively, compared with approximately 24M in the full setting. Samples removed from the paired pool are excluded from training rather than converted to unpaired samples.

Dataset-level mixture probabilities are kept unchanged for these nonzero settings. Once a paired dataset is selected, samples are uniformly drawn from its reduced active index pool. Consequently, retained paired demonstrations are revisited more frequently, while the expected paired-supervision exposure remains approximately fixed.

For the 0\% setting, we remove all paired datasets and train solely on the original unpaired datasets. Thus, cross-prediction and pair-alignment objectives requiring paired samples are inactive, while self-prediction and auxiliary supervision remain available. This setting serves as an unpaired-only reference rather than a fixed-paired-exposure condition.

The original unpaired datasets, preprocessing, evaluation sets, and total optimization budget are kept unchanged. Visual and latent-action retrieval are evaluated for all five settings using the same held-out paired evaluation pool as Sec.~\ref{sec:alignment}. RLBench and CALVIN are evaluated for the 100\%, 25\%, and 5\% settings following the same downstream protocols as the full-data model.

\paragraph{Controlled comparison.}
For controlled HARP variants, all methods use the same paired/unpaired training split, DTW preprocessing, auxiliary cues, visual backbone, robot-only adapter, and training budget; only the alignment objective or pooling space differs.
HR-Align baselines use the same fused backbone and robot-only adapter but follow the original paired-data alignment setup.

%% file: Appendix/rlbench_eval.tex
We follow the RLBench evaluation protocol used by HR-Align\cite{Zhou2024MitigatingTH} and evaluate frozen visual encoders on 18 RLBench manipulation tasks.
To isolate the effect of visual representation quality, all methods use the same downstream policy architecture, policy head, action space, demonstration data, training schedule, and optimization budget.
The only component changed across methods is the frozen visual encoder used to extract visual representations.

For each method, the corresponding visual encoder is initialized from the evaluated representation-learning checkpoint and kept frozen during downstream policy learning.
The policy head receives frozen visual features and predicts robot actions under the same imitation-learning objective for all methods.
Each task is evaluated over 75 episodes using the standard RLBench success signal, resulting in 1,350 evaluation episodes per method.
We report the average success rate across the 18 tasks in Table~\ref{tab:rlbench_repr}.

\newcommand{\taskname}[1]{\texttt{\detokenize{#1}}}

The evaluated RLBench tasks are:
\taskname{put_item_in_drawer}, \taskname{reach_and_drag}, \taskname{turn_tap}, \taskname{slide_block_to_color_target}, \taskname{open_drawer}, \taskname{put_groceries_in_cupboard},
\taskname{place_shape_in_shape_sorter}, \taskname{put_money_in_safe}, \taskname{push_buttons}, \taskname{close_jar}, \taskname{stack_blocks}, \taskname{place_cups},
\taskname{place_wine_at_rack_location}, \taskname{light_bulb_in}, \taskname{sweep_to_dustpan_of_size}, \taskname{insert_onto_square_peg}, \taskname{meat_off_grill}, and \taskname{stack_cups}.